\documentclass[border=11pt]{article}
\usepackage[preprint]{acl}

\usepackage{placeins}
\usepackage{graphicx}
\usepackage{booktabs}
\usepackage{siunitx}
\usepackage{times}
\usepackage{latexsym}
\usepackage{tikz}
\usepackage{amssymb}
\usepackage[table]{xcolor}
\usepackage{subcaption}
\DeclareUnicodeCharacter{202F}{~}

\usetikzlibrary{shapes.geometric, arrows.meta, positioning, calc}

\usepackage[T1]{fontenc}

\usepackage[utf8]{inputenc}

\usepackage{microtype}

\usepackage{inconsolata}

\usepackage{graphicx}

\title{Polish ModernBERT: The Long and Short of Polish Language Understanding}

\author{
Michał Perełkiewicz$^{*}$, Sławomir Dadas, Rafał Poświata, Małgorzata Grębowiec \\
National Information Processing Institute, Warsaw, Poland \\
\texttt{\{mperelkiewicz,sdadas,rposwiata,mgrebowiec\}@opi.org.pl} \\
$^{*}$Corresponding author
}

\begin{document}
\maketitle

\begin{abstract}
Encoder-only Transformers remain effective for discriminative and
representation-learning tasks, yet Polish encoders still largely rely on
BERT/RoBERTa-style architectures. We introduce \textbf{Polish ModernBERT}, a family of four Polish encoders available at Base and Large scales, each
with 512-token and 8K context variants. We adapt the ModernBERT pretraining recipe through
staged selection experiments and release a long-context benchmark covering legal topic classification, ideological decision-direction
prediction, factual-consistency assessment over literary plot summaries,
and human-rights violation assessment. Across 30 tasks, Polish ModernBERT achieves the best
overall performance among the evaluated Polish encoders, reaching 83.99 and
85.11 for the Base-8K and Large-8K models, respectively. On long-context
tasks, the 8K variants improve over matched Polish RoBERTa-8K baselines from 67.47 to 77.15 and from 75.88 to 78.49 at the Base and Large scales,
respectively. The Base-8K model achieves this gain with 22\% fewer parameters
(149M vs.\ 190M). Efficiency measurements in representative inference setups show lower peak
memory usage and latency than matched Polish RoBERTa baselines in both
512-token and 8K settings. Polish ModernBERT-8K-Base additionally achieves
the best result on a Polish retrieval benchmark among the evaluated encoders
below 300M parameters.
\end{abstract}

\section{Introduction}

Recent progress in NLP has been driven largely by decoder-only large language
models. Encoder-only Transformers, however, remain a competitive and computationally efficient choice for
discriminative and representation-learning applications, including text
classification, dense retrieval, cross-encoder reranking, and sentence representation learning. Recent architectures such as MosaicBERT
\citep{NEURIPS2023_095a6917}, ModernBERT
\citep{warner-etal-2025-smarter}, and NeoBERT
\citep{breton2025neobertnextgenerationbert} demonstrate that BERT-style
models can be further improved through updated architectural choices,
training recipes, and optimized implementations.

Access to these advances remains uneven across languages. Although
ModernBERT-style architectures have recently been extended to several
languages and multilingual settings
\citep{pellicer-rinaldo-2026-norberto,
shmidman2025neodictabertpushingfrontierbert,
marone2025mmbertmodernmultilingualencoder,
boizard2025eurobertscalingmultilingualencoders},
many languages continue to rely primarily on earlier BERT
\citep{devlin-etal-2019-bert} and RoBERTa \citep{liu2019roberta}
architectures. This limitation is particularly relevant for morphologically
rich languages such as Polish, for which language-specific tokenization and
monolingual pretraining can provide better lexical coverage and stronger
downstream representations.

Existing Polish encoders include HerBERT
\citep{mroczkowski-etal-2021-herbert}, Polish RoBERTa-v2
\citep{dadas2020pre}, and Polish RoBERTa-8K
\citep{dadas2026longcontext}. Although these models remain strong reference
points, to our knowledge, no previous Polish encoder family combines a
ModernBERT-based architecture with both Base and Large model scales and
512-token and 8K context lengths.

To address this gap, we introduce \textbf{Polish ModernBERT}\footnote{\url{https://huggingface.co/collections/OPI-PIB/pl-modernbert}}, a family of
four Polish encoder models covering Base and Large scales and maximum context
lengths of 512 and 8,192 tokens. We adapt the ModernBERT pretraining recipe to
Polish through staged recipe selection, corpus refinement, and long-context
continuation. We additionally introduce the \textbf{LongContext benchmark}\footnote{\url{https://huggingface.co/collections/mmichall/polish-longcontext-benchmark}}, a
five-task Polish evaluation suite for long-document understanding.

Across 30 downstream tasks, Polish ModernBERT achieves stronger aggregate performance than prior Polish encoders, with particularly strong gains on long-document
understanding. The Base-8K model improves over the matched Polish RoBERTa-8K
baseline by 9.68 points while using 22\% fewer parameters, and the Large-8K
variant achieves the best overall score among the evaluated Polish encoders.
The models also reduce inference latency and peak memory usage relative to
matched RoBERTa baselines and achieve strong performance on a Polish retrieval
benchmark.

Our contributions are as follows:
(i) we introduce a ModernBERT-based encoder family for Polish, covering Base
and Large scales and 512-token and 8K context lengths, and will release all
four checkpoints publicly upon publication;
(ii) we adapt the ModernBERT pretraining setup through a staged procedure
encompassing recipe selection, corpus refinement, and long-context
continuation;
(iii) we introduce LongContext, a five-task Polish benchmark specifically
designed to evaluate long-document understanding beyond the standard
512-token context window, covering legal topic and decision-direction
prediction, human-rights violation assessment, and factual-consistency
assessment over literary plot summaries; and
(iv) we provide an extensive evaluation against Polish and multilingual
encoders and retrieval models, together with inference-efficiency
measurements showing favorable quality--cost trade-offs.

\section{Related Work}

BERT \citep{devlin-etal-2019-bert} established encoder-only Transformers as
a central architecture for discriminative NLP. Subsequent models improved
different aspects of the original framework: RoBERTa
\citep{liu2019roberta} demonstrated the importance of data scale and an
optimized pretraining recipe, ELECTRA \citep{clark2020electra} introduced
replaced-token detection as a more sample-efficient alternative to masked
language modeling, and DeBERTa \citep{DBLP:journals/corr/abs-2006-03654}
proposed disentangled attention and enhanced position representations.

Long-document encoders such as Longformer
\citep{Beltagy2020LongformerTL} and BigBird
\citep{3495724.3497174} extended the context length of BERT-style models
through sparse attention mechanisms. More recent work has revisited the
broader encoder design by incorporating architectural, training, and
implementation advances developed for later Transformer systems.
MosaicBERT \citep{NEURIPS2023_095a6917} focuses on efficient pretraining,
while ModernBERT \citep{warner-etal-2025-smarter} and NeoBERT
\citep{breton2025neobertnextgenerationbert} combine updated architectural
choices with optimized implementations and native support for contexts of
up to 8,192 and 4,096 tokens, respectively.

Multilingual encoders such as mBERT
\citep{devlin-etal-2019-bert,pires-etal-2019-multilingual} and XLM-R
\citep{conneau2019unsupervised} established strong baselines for
cross-lingual transfer. More recently, multilingual encoders have also been
revisited using modern architectural and training choices. EuroBERT
\citep{boizard2025eurobertscalingmultilingualencoders} introduces a family
of encoders covering European and widely spoken global languages with
native 8K-token support, while mmBERT
\citep{marone2025mmbertmodernmultilingualencoder} scales modern encoder
pretraining to a massively multilingual setting. Nevertheless, sharing a
fixed model capacity across many languages can limit performance for
individual languages, particularly when they are morphologically rich or
underrepresented in the pretraining mixture
\citep{conneau2019unsupervised,hu2020xtrememassivelymultilingualmultitask,
ruder-etal-2021-xtreme}.

This limitation has motivated complementary work on language-focused modern
encoders, including Modern-LiBERTa for Ukrainian
\citep{haltiuk-smywinski-pohl-2025-path}, NeoDictaBERT for Hebrew
\citep{shmidman2025neodictabertpushingfrontierbert}, Finnish-centered
ModernBERT models \citep{reunamo2025pretrainingfinnishmodernberts},
TabiBERT for Turkish \citep{Trker2025TabiBERTAL}, and NorBERTo for Brazilian
Portuguese \citep{pellicer-rinaldo-2026-norberto}.

For Polish, prominent general-purpose encoder families include HerBERT
\citep{mroczkowski-etal-2021-herbert} and Polish RoBERTa
\citep{dadas2020pre}. Polish RoBERTa-8K
\citep{dadas2026longcontext} further extends this line of work to
long-document processing through continued pretraining of a RoBERTa-based
model. Despite these developments, Polish still lacks a modern encoder
family covering multiple model scales and both standard- and long-context
settings.

\section{Polish ModernBERT}
\label{sec:polish_modernbert}

\subsection{Model Family and Architecture}
\label{sec:architecture}

All Polish ModernBERT variants retain the core ModernBERT architecture
\citep{warner-etal-2025-smarter}, including rotary positional embeddings,
GeGLU feed-forward layers, pre-normalization, and alternating global and local
sliding-window attention.

The model family varies primarily along two axes: model capacity and context
length, with tokenizer design coupled to model scale. We release Base and Large
models with maximum context lengths of 512 and 8,192 tokens. The 512-token
models are pretrained from scratch on Polish data. The corresponding 8K
variants are initialized from these checkpoints and undergo continued
pretraining for long-context processing. During context extension, the global
RoPE $\theta$ is increased from $10{,}000$ to $160{,}000$, while the local
RoPE configuration remains unchanged.

The 512-token variants target conventional short-text processing and
lower-cost inference, whereas the 8K variants are intended for document-level
tasks. Within each model scale, the two context variants share the same
architecture, tokenizer, and parameter count. Table~\ref{tab:model_variants}
summarizes their configurations.

\begin{table}[!th]
\centering
\small
\setlength{\tabcolsep}{3.5pt}
\caption{
Polish ModernBERT configurations. Slash-separated values denote the 512- and 8K-token variants, respectively.
}
\label{tab:model_variants}
\begin{tabular}{lcc}
\toprule
\textbf{Property} & \textbf{Base} & \textbf{Large} \\
\midrule
Context              & 512 / 8,192 & 512 / 8,192 \\
Vocabulary           & 50,008 & 128,256 \\
Byte fallback        & No & Yes \\
\midrule
Layers               & 22 & 28 \\
Hidden size          & 768 & 1,024 \\
FFN size             & 1,152 & 2,624 \\
GLU size             & 2,304 & 5,248 \\
Attention heads      & 12 & 16 \\
Global attention     & Every 3rd & Every 3rd \\
Local window         & 128 & 128 \\
Global RoPE $\theta$ & $10{,}000 / 160{,}000$ & $10{,}000 / 160{,}000$ \\
Local RoPE $\theta$  & $10{,}000$ & $10{,}000$ \\
\midrule
Parameters           & 149M & 475M \\
Non-embedd. params & 111M & 343M \\
\bottomrule
\end{tabular}
\vspace{-0.2cm}
\end{table}

\paragraph{Tokenization}
\label{sec:tokenization}

The Base variants use the 50K SentencePiece Unigram tokenizer adopted from
Polish RoBERTa-v2 \citep{dadas2020pre}. We extend its original vocabulary from
50,001 to 50,008 entries by adding seven unused tokens, making the vocabulary
size divisible by eight and aligning the embedding matrix with
hardware-friendly tensor dimensions. As shown in
Table~\ref{tab:dataset_length_stats}, the tokenizer provides broad coverage
across most evaluation datasets, with higher unknown-token rates concentrated
primarily in noisy user-generated and social-media text.

For the Large variants, we train a new SentencePiece Unigram tokenizer
\citep{kudo-richardson-2018-sentencepiece} with a total vocabulary of 128,256
entries, including byte-level fallback symbols. The larger vocabulary is
designed to provide finer segmentation of rare, domain-specific, and
morphologically complex Polish word forms, while byte fallback guarantees that
arbitrary input strings can be represented without unknown tokens. We use this
tokenizer only for the Large variants, for which the additional embedding
parameters constitute a smaller proportion of the overall model capacity.

\subsection{Pretraining Corpus}
\label{sec:pretraining_corpus}

The main Polish ModernBERT pretraining corpus comprises 44.5B tokens from
three sources: a curated Polish corpus, a Polish subset of Common Crawl, and
a filtered Polish subset of FineTranslations
\citep{penedo2026finetranslations}. Table~\ref{tab:pretraining_data}
summarizes the post-processing size of each component, while a domain-level
breakdown is provided in Appendix~\ref{app:domain_composition}.

The curated component contains Polish-language documents collected from
publicly available web sources and existing text collections. It covers
encyclopedic, scientific, educational, legal, parliamentary, literary,
question--answer, consumer-review, news, discussion-forum, and general web
content.

We include Polish documents extracted from the October 2019 Common Crawl
snapshot (\texttt{CC-MAIN-2019-43}) to broaden the coverage of web language
and long-tail lexical phenomena.

We also use Polish text from the \texttt{pol\_Latn} subset of
FineTranslations, a multilingual dataset derived from FineWeb2 and released
by Hugging Face.\footnote{\url{https://huggingface.co/datasets/HuggingFaceFW/finetranslations}}
We retain documents with \texttt{edu\_score\_raw > 1.24}, where
\texttt{edu\_score\_raw} is a document-level educational-quality score
provided with the dataset, with higher values indicating higher estimated
quality. The threshold corresponds to approximately the top 25\% of the Polish
subset. To increase the representation of long-form content, we additionally
retain documents longer than 2,000 tokens regardless of this score. These
documents remain subject to the shared cleaning and quality-filtering pipeline.
Documents admitted through this length-based criterion account for
approximately 27\% of the retained FineTranslations component, or 3.3B of
12.4B tokens.

\paragraph{Stage-specific mixtures.}
In addition to the main pretraining corpus, we construct two mixtures for
specific stages of training. The \textbf{annealing corpus} is derived from
the curated Polish corpus by upsampling selected components, including
Wikipedia, legal texts, and consumer reviews. It is used during the final
512-token pretraining stage to increase the proportion of encyclopedic,
legal, and opinion-oriented content. The \textbf{context-extension corpus} combines the curated Polish corpus with
the FineTranslations component, including the additionally retained long
documents. It is used for continued pretraining with sequences of up to
8,192 tokens and increases the representation of long-form content during
context extension. Both mixtures are constructed from the cleaned and
deduplicated source corpora and therefore do not introduce additional unique
pretraining data.

\paragraph{Corpus cleaning.}
Following source-specific extraction, all corpus components undergo a shared
cleaning and filtering pipeline comprising sentence segmentation, punctuation
and whitespace normalization, URL removal, sentence-level language
identification, heuristic and classifier-based quality filtering, and
KenLM-based perplexity filtering. Documents shorter than 500 characters are
removed. We subsequently apply exact and near-duplicate removal across the
corpus components.

After cleaning and deduplication, approximately 0.1\% of documents are reserved
as a held-out masked-language-modeling validation set and excluded from all
pretraining stages. Full preprocessing and deduplication details are provided
in Appendix~\ref{app:corpus_cleaning}.

\begin{table}[t]
\centering
\small
\setlength{\tabcolsep}{4pt}
\caption{
Post-cleaning pretraining-corpus statistics. Token counts use the Base
tokenizer; annealing statistics reflect the upsampled mixture.
}
\label{tab:pretraining_data}
\begin{tabular}{lrr}
\toprule
\textbf{Corpus / mixture} & \textbf{Text size} & \textbf{Tokens} \\
\midrule
\multicolumn{3}{l}{\textbf{Main pretraining corpus}} \\
\midrule
Curated Polish corpus       & 70 GB  & 15.8B \\
Common Crawl                & 70 GB  & 16.3B \\
FineTranslations            & 57 GB  & 12.4B \\
\midrule
Total main pretraining data & 197 GB & 44.5B \\
\midrule
\multicolumn{3}{l}{\textbf{Stage-specific mixtures}} \\
\midrule
Annealing corpus            & 86 GB  & 18.6B \\
Context-extension corpus    & 127 GB & 28.2B \\
\bottomrule
\end{tabular}
\vspace{-0.2cm}
\end{table}

\subsection{Training Procedure}
\label{sec:training_procedure}

\begin{figure*}[t]
\centering
\includegraphics[width=\textwidth,height=6.6cm]{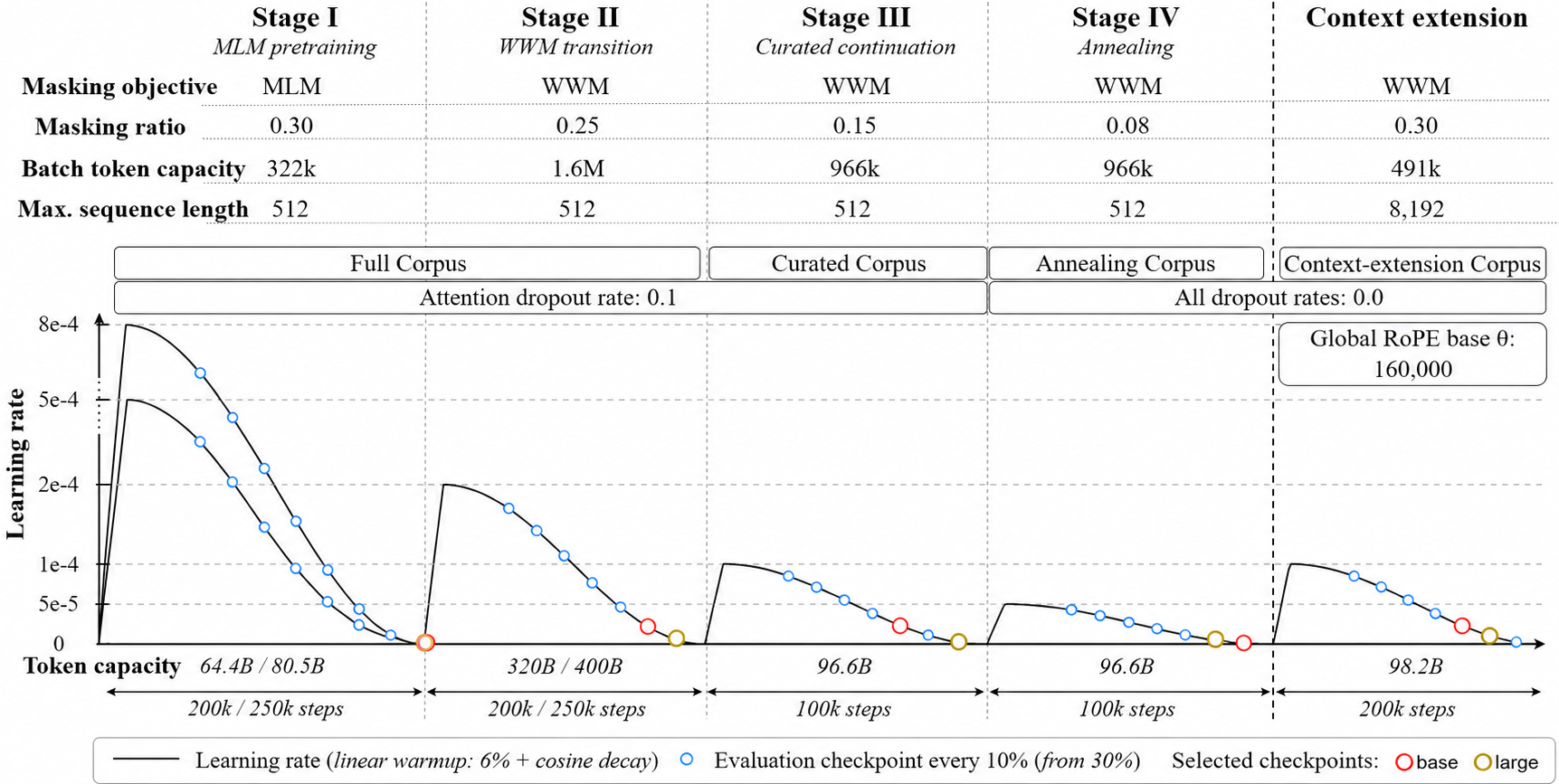}
\caption{
Polish ModernBERT pretraining schedule. The models are trained in four
512-token stages followed by 8K continuation. Blue markers denote evaluated
checkpoints; red/gold markers denote selected Base/Large checkpoints.
Slash-separated values correspond to Base / Large variants.
}
\vspace{-0.2cm}
\label{fig:training_schedule}
\end{figure*}

The final pretraining schedule was selected through a compute-aware staged
procedure initialized from the original ModernBERT recipe
\citep{warner-etal-2025-smarter}. As an initial baseline, we directly adapted
the original configuration to Polish. This baseline achieved lower average
KLEJ validation performance than the configurations identified through the
subsequent recipe search; its training setup and downstream results are
reported in Appendix~\ref{app:org_recipe}.

Recipe selection was conducted on the Base model. At each stage, we varied a
small subset of factors---the learning-rate schedule, masking objective and
ratio, corpus mixture, or peak learning rate---while keeping the remaining
configuration fixed. Intermediate checkpoints were evaluated using the average
KLEJ validation score, and the best-performing checkpoint was propagated to
the subsequent stage. The selected recipe was then transferred to the Large
variants. Following the original ModernBERT settings, Stage~I used peak
learning rates of $8\times10^{-4}$ and $5\times10^{-4}$ for Base and Large,
respectively. For Large, Stages~I--II were each extended by 50K steps to
accommodate the greater model capacity and vocabulary size. The evaluated
configurations and validation results are provided in
Appendix~\ref{app:pretraining_recipe_selection}.

\paragraph{Overall schedule.}
Figure~\ref{fig:training_schedule} summarizes the resulting training
trajectory. The models undergo four 512-token pretraining stages, followed by
long-context continuation from the selected Stage~IV checkpoints. Across the
512-token stages, training progresses from token-level MLM on the full corpus
to whole-word masking on the full, curated, and annealing corpora, with
progressively lower masking ratios and peak learning rates. The final
512-token checkpoints initialize the corresponding 8K variants, which undergo
continued pretraining on the context-extension corpus after increasing the
global RoPE base from 10,000 to 160,000.

Each stage uses dynamic masking and its own learning-rate schedule, comprising
6\% linear warm-up followed by cosine decay
\citep{Loshchilov2016SGDRSG}. At each stage transition, the optimizer and
scheduler states are reinitialized, yielding stage-wise cosine restarts aligned
with changes in the masking objective, masking ratio, or corpus mixture.

\paragraph{Token accounting.}
The token budgets shown in Figure~\ref{fig:training_schedule} denote nominal
capacity before excluding padding tokens. With sequence unpadding, non-padding
tokens account for approximately 56\% of this capacity during 512-token
pretraining and 16\% during context extension. Along the trajectories leading
to the selected checkpoints, the Base and Large models process approximately
271.4B and 344.1B non-padding tokens, respectively, during 512-token
pretraining, followed by approximately 11.0B and 12.6B tokens during context
extension.

\paragraph{Implementation and compute.}
We use the official ModernBERT implementation released by the original
authors\footnote{\url{https://github.com/answerdotai/modernbert}}, including
sequence unpadding, optimized attention kernels, and Megatron-style parameter
initialization. Training is performed with Composer 0.30.0 using distributed
data parallelism across eight NVIDIA GH200 GPUs on two HPC nodes. We optimize
all models with decoupled StableAdamW
\citep{wortsman2023stable,Loshchilov2017DecoupledWD} and use BF16 mixed
precision throughout pretraining. 

Full optimizer settings,
checkpoint-selection criteria, stage-wise hyperparameters, and token budgets
are provided in Appendix~\ref{app:pretraining_recipe_selection}.

\section{Evaluation}
\label{sec:evaluation}


The evaluation suite comprises 30 Polish NLU tasks organized into four groups:
nine KLEJ tasks, six FinBench tasks, ten additional tasks grouped under
\textit{Other Tasks}, and the five-task \textbf{LongContext} benchmark.
KLEJ was introduced by \citet{rybak-etal-2020-klej}, whereas FinBench and the
original nine-task \textit{Other Tasks} group were introduced by
\citet{dadas2026longcontext}. In this work, we extend the latter group with
one additional task and introduce the LongContext benchmark.

KLEJ covers sentiment analysis (\textsc{polemo2.0-in},
\textsc{polemo2.0-out}, and \textsc{ar}), named-entity type classification
(\textsc{nkjp-ner}), harmful-content detection (\textsc{cbd}), and
semantic-relation tasks (\textsc{cdsc-e}, \textsc{cdsc-r}, \textsc{dyk},
and \textsc{psc}). FinBench evaluates financial-domain understanding through
topic classification of short and long banking texts
(\textsc{banking-short} and \textsc{banking-long}), intent detection
(\textsc{banking77}), sentiment analysis (\textsc{fpb} and \textsc{stooq}),
and multi-label topic classification (\textsc{gcn}). The ten additional tasks
grouped under \textit{Other Tasks} broaden the evaluation to semantic
relations (\textsc{ppc}, \textsc{sick-e}, and \textsc{sick-r}), thematic
classification (\textsc{8tags} and \textsc{eurlex}), sentiment and emotion
analysis (\textsc{imdb} and \textsc{twitteremo}), harmful-content and
manipulation detection (\textsc{ban-pl} and \textsc{mipd}), and sequence
labeling (\textsc{nkjp-ner*}).

LongContext specifically targets document understanding beyond the standard
512-token context window. It comprises tasks selected or constructed to
evaluate long-document processing and distinguish between standard- and
extended-context encoders.

\paragraph{NKJP-NER*.}
\textsc{nkjp-ner*} is a sequence-labeling task derived from the Polish
National Corpus and included among Other Tasks. Unlike the KLEJ
\textsc{nkjp-ner} formulation, which recasts the data as single-label
classification, it preserves the original sequence-labeling setup\footnote{
The processed \textsc{nkjp-ner*} dataset is available at
\url{https://anonymous.4open.science/r/pl-modernbert-5E28}.
}. Using the
same underlying texts and annotations, we evaluate 512-token models on
sentence-level inputs and 8K models on full documents. Both variants are
treated as one task and evaluated using micro-averaged entity-level F1. This
setup evaluates each model under its intended input regime and should not be
interpreted as a controlled comparison of context length alone.

The complete list of tasks, prediction formulations, domains, and primary
metrics is provided in Table~\ref{tab:evaluation_tasks}. Test-set sizes and
token-length statistics are reported in
Table~\ref{tab:dataset_length_stats}.

\begin{table*}[t]
\vspace{-0.1cm}
\centering
\small

\caption{
Base-scale encoder results across 30 tasks. Boldface marks the best score in
each row, while underlining marks the best score within each context-length
group. Averages are unweighted macro-averages across tasks. Superscripts
denote architecture families: $^{\mathrm{B}}$BERT,
$^{\mathrm{R}}$RoBERTa, $^{\mathrm{M}}$ModernBERT, and
$^{\mathrm{L}}$Llama-inspired encoder; (m) denotes multilingual models.
POLEMO-IN/OUT denote the in- and out-of-domain POLEMO2.0 variants.
NKJP-NER* uses sentence-level inputs for 512-token models and document-level
inputs for 8K models.
}
\label{tab:base_results}

\resizebox{\textwidth}{!}{
\begin{tabular}{l S S S S !{\color{gray!60}\vrule width 0.3pt} S S S S }
\toprule
&
\multicolumn{4}{c}{\textbf{512 tokens}} &
\multicolumn{4}{c}{\textbf{8K tokens}} \\
\cmidrule(lr){2-5} \cmidrule(lr){6-9}

& {\rotatebox{0}{\begin{tabular}{@{}c@{}}
XLM-RoBERTa$^{\mathrm{R}}$(m) \\
\textit{base} (278M)
\end{tabular}}}
& {\rotatebox{0}{\begin{tabular}{@{}c@{}}
HerBERT$^{\mathrm{B}}$ \\
\textit{base} (124M)
\end{tabular}}} 
& {\rotatebox{0}{\begin{tabular}{@{}c@{}}
pl-RoBERTa-v2$^{\mathrm{R}}$ \\
\textit{base} (124M)
\end{tabular}}}
& {\rotatebox{0}{\begin{tabular}{@{}c@{}}
\textbf{pl-ModernBERT}$^{\mathrm{M}}$ \\
\textit{base} (149M)
\end{tabular}}}
& {\rotatebox{0}{\begin{tabular}{@{}c@{}}
EuroBERT$^{\mathrm{L}}$(m) \\
\textit{210m} (212M)
\end{tabular}}}
& {\rotatebox{0}{\begin{tabular}{@{}c@{}}
mmBERT$^{\mathrm{M}}$(m) \\
\textit{small} (140M)
\end{tabular}}}
& {\rotatebox{0}{\begin{tabular}{@{}c@{}}
pl-RoBERTa-8K$^{\mathrm{R}}$ \\
\textit{base} (190M)
\end{tabular}}}
& {\rotatebox{0}{\begin{tabular}{@{}c@{}}
\textbf{pl-ModernBERT-8K}$^{\mathrm{M}}$ \\
\textit{base} (149M)
\end{tabular}}} \\

\midrule
\multicolumn{9}{l}{\textbf{KLEJ}} \\
\midrule
                                                                                             
NKJP-NER    & 92.37 & 94.13 & 94.32 & \underline{94.38} & 84.17 & 90.99 & 94.16 & \textbf{\underline{94.49}}   \\
CDSC-E      & 93.50 & 94.22 & 94.05 & \textbf{\underline{94.66}} & 92.64 & 93.34 & \underline{94.54} & 94.46 \\
CDSC-R      & 93.38 & 93.84 & \underline{94.64} & 94.11 & 88.40 & 90.47 & \textbf{\underline{94.90}} & 94.06 \\
CBD         & 60.84 & 66.36 & \underline{70.57} & 68.56 & 51.06 & 50.02 & 69.35 & \textbf{\underline{71.40}} \\
POLEMO-IN   & 90.80 & 90.50 & 90.97 & \textbf{\underline{92.88}} & 88.28 & 89.36 & 91.27 & \underline{92.14} \\
POLEMO-OUT  & 79.27 & 77.94 & 79.11 & \textbf{\underline{83.77}} & 73.44 & 76.88 & 81.26 & \underline{83.04} \\
DYK         & 65.02 & 68.82 & \textbf{\underline{70.38}} & 66.90 & 36.41 & 48.28 & \underline{69.35} & 67.28  \\
PSC         & 97.91 & \textbf{\underline{98.94}} & 98.88 & 97.79 & 95.30 & 97.10 & \underline{98.90} & 97.68 \\
AR          & 87.06 & 87.74 & 87.83 & \textbf{\underline{88.53}} & 84.73 & 84.59 & 88.05 & \underline{88.46} \\

\rowcolor{gray!9}
\textit{Average}     & 84.46 & 85.83 & 86.75 & \underline{86.84} & 77.16 & 80.11 & 86.86 & \textbf{\underline{87.00}} \\

\midrule
\multicolumn{9}{l}{\textbf{FinBench}} \\
\midrule

Banking-Short  & 77.34& 78.35 & 78.75 & \textbf{\underline{80.41}} & 73.72 & 73.86 & 79.79 & \underline{80.08} \\
Banking-Long   & 82.65& 85.09 & 85.03 & \textbf{\underline{87.29}} & 86.34 & 85.13 & 86.99 & \underline{87.16} \\
Banking77      & 85.10& 87.29 & 88.26 & \textbf{\underline{91.85}} & 91.15 & 88.97 & 89.27 & \underline{91.66} \\
FPB            & 82.76& 83.11 & \underline{83.55} & 83.20 & 78.52 & 81.77 & \textbf{\underline{83.63}} & 83.40 \\
GCN            & 94.38& 94.73 & \textbf{\underline{95.02}} & 94.87 & 94.71 & 94.61 & \underline{94.87} & 94.83 \\
Stooq          & 76.53& 73.33 & 80.25 & \textbf{\underline{84.08}} & 69.81 & 74.05 & 81.32 & \underline{83.03} \\

\rowcolor{gray!9}
\textit{Average}  & 83.13 & 83.65 & 85.14 & \textbf{\underline{86.95}} & 82.38 & 83.07 & 85.98 & \underline{86.69} \\

\midrule
\multicolumn{9}{l}{\textbf{Other Tasks}} \\
\midrule

8TAGS      & 76.15 & 77.81 & 78.03 & \underline{80.69} & 72.17 & 74.41 & 79.21 & \textbf{\underline{80.86}} \\
BAN-PL     & 90.71 & 91.71 & 92.19 & \textbf{\underline{93.10}}  & 88.89 & 90.37 & 92.62 & \underline{93.08} \\
MIPD       & 54.30 & 57.65 & 58.58 & \underline{67.11} & 64.96 & 66.95 & 64.39 & \textbf{\underline{68.03}} \\
PPC        & 81.76 & 84.16 & \textbf{\underline{87.05}} & 84.40 & 66.54 & 80.10 & \underline{86.02} & 85.70 \\
SICK-E     & 83.98 & 85.17 & \textbf{\underline{86.71}} & 86.31 & 75.34 & 84.35 & 86.31 & \underline{86.61} \\
SICK-R     & 73.79 & 77.82 & 82.58 & \underline{83.16} & 52.13 & 66.38 & 83.16 & \textbf{\underline{83.77}} \\
TwitterEMO  & 65.19 & 67.41 & 66.46 & \underline{69.02} & 61.93 & 63.23 & 68.75 & \textbf{\underline{69.52}} \\
IMDB       & 88.12 & 90.21 & 91.06 & \underline{92.05} & 91.36 & 92.10 & \textbf{\underline{95.02}} & 94.40 \\
EURLEX     & 72.60 & 75.37 & 74.51 & \underline{79.31} & 79.00 & 78.36 & 79.12 & \textbf{\underline{79.61}} \\
NKJP-NER*           & 84.36 & 85.49 & 85.41 & \underline{85.54} & 84.76 & 83.82 & 88.97 & \textbf{\underline{89.21}} \\ 

\rowcolor{gray!9}
\textit{Average}  & 77.10 & 79.28 & 80.26 & \underline{82.07} & 73.71 & 78.01 & 82.36 & \textbf{\underline{83.08}}
\\

\midrule
\rowcolor{gray!11}
\textbf{Short Ctx Avg}  & 81.19 & 82.69 & 83.77 & \underline{84.96} & 77.03 & 79.98 & 84.85 & \textbf{\underline{85.36}}
\\ 

\midrule
\multicolumn{9}{l}{\textbf{LongContext}} \\
\midrule

SCOTUS-Dom              & 75.36 & 78.80 & 79.12 & \underline{82.81} & 83.45 & 79.83 & 79.26 & \textbf{\underline{84.48}} \\
SCOTUS-Dec      & 61.46 & 65.59 & 69.86 & \underline{70.74} & 76.23 & 60.98 & 63.20 & \textbf{\underline{77.79}} \\
BookSummary            & 78.53 & 81.12 & \underline{85.02} & 83.71 & 82.27 & 82.58 & 88.96 & \textbf{\underline{90.22}} \\ 
ECtHR-PL-AVA        & 22.04 & 43.90 & 33.65 & \underline{61.42} & 62.47 & 60.26 & 64.66 & \textbf{\underline{68.01}} \\
ECtHR-PL-VA         & 16.38 & 40.26 & 20.65 & \underline{59.29} & 57.09 & 61.45 & 41.28 & \textbf{\underline{65.27}}\\

\midrule
\rowcolor{gray!11}
\textbf{Long Ctx Avg}  & 50.75 & 61.93 & 57.66 & \underline{71.59} & 72.30 & 69.02 & 67.47 & \textbf{\underline{77.15}}
\\ 

\midrule
\rowcolor{gray!11}
\textbf{Overall Average}  & 76.12 & 79.23 & 79.42 & \underline{82.73} & 76.24 & 78.15 & 81.95 & \textbf{\underline{83.99}} \\

\bottomrule
\end{tabular}
}

\vspace{-0.2cm}
\end{table*}

\begin{table*}[t]
\vspace{-0.1cm}
\centering
\small
\caption{
Large-scale encoder results across 30 tasks. Formatting and notation follow
Table~\ref{tab:base_results}.
}
\label{tab:large_results}
\resizebox{\textwidth}{!}{
\begin{tabular}{l S S S S !{\color{gray!60}\vrule width 0.3pt} S S S S }
\toprule
&
\multicolumn{4}{c}{\textbf{512 tokens}} &
\multicolumn{4}{c}{\textbf{8K tokens}} \\
\cmidrule(lr){2-5} \cmidrule(lr){6-9}

& {\rotatebox{0}{\begin{tabular}{@{}c@{}}
XLM-RoBERTa$^{\mathrm{R}}$(m) \\
\textit{large} (560M)
\end{tabular}}}
& {\rotatebox{0}{\begin{tabular}{@{}c@{}}
HerBERT$^{\mathrm{B}}$ \\
\textit{large} (355M)
\end{tabular}}} 
& {\rotatebox{0}{\begin{tabular}{@{}c@{}}
pl-RoBERTa-v2$^{\mathrm{R}}$ \\
\textit{large} (435M)
\end{tabular}}}
& {\rotatebox{0}{\begin{tabular}{@{}c@{}}
\textbf{pl-ModernBERT}$^{\mathrm{M}}$ \\
\textit{large} (475M)
\end{tabular}}}
& {\rotatebox{0}{\begin{tabular}{@{}c@{}}
EuroBERT$^{\mathrm{L}}$(m) \\
\textit{610m} (610M)
\end{tabular}}}
& {\rotatebox{0}{\begin{tabular}{@{}c@{}}
mmBERT$^{\mathrm{M}}$(m) \\
\textit{base} (307M)
\end{tabular}}}
& {\rotatebox{0}{\begin{tabular}{@{}c@{}}
pl-RoBERTa-8K$^{\mathrm{R}}$ \\
\textit{large} (443M)
\end{tabular}}}
& {\rotatebox{0}{\begin{tabular}{@{}c@{}}
\textbf{pl-ModernBERT-8K}$^{\mathrm{M}}$ \\
\textit{large} (475M)
\end{tabular}}} \\

\midrule
\multicolumn{9}{l}{\textbf{KLEJ}} \\
\midrule

NKJP-NER    & 94.68 & \underline{\textbf{96.07}} & 95.75 & 95.05 & 89.59 & 92.34 & \underline{95.64} & 94.38  \\
CDSC-E      & 94.40 & \underline{\textbf{94.78}} & 94.16 & 94.60 & 89.48 & 93.72 & 94.28 & \underline{94.68}  \\
CDSC-R      & 94.75 & 95.01 & \underline{95.25} & 95.14 & 91.14 & 93.12 & \underline{\textbf{95.33}} & 94.47  \\
CBD         & 66.91 & 70.21 & \underline{73.10} & 72.60 & 54.58 & 52.55 & \underline{\textbf{73.23}} & 71.47  \\
POLEMO-IN   & 92.47 & 91.39 & \underline{\textbf{93.55}} & 93.38 & 90.89 & 90.22 & \underline{93.05} & \underline{93.05}  \\
POLEMO-OUT  & 81.78 & 81.66 & 83.81 & \underline{84.41} & 78.42 & 78.38 & 83.64 & \underline{\textbf{84.78}}  \\
DYK         & 73.16 & 73.31 & \underline{\textbf{74.87}} & 73.28 & 42.61 & 64.24 & 74.05 & \underline{74.63}  \\
PSC         & \textbf{\underline{98.91}} & 98.85 & 98.37 & 98.81 & 98.03 & 96.83 & \underline{98.56} & 98.47 \\
AR          & 88.53 & 89.23 & \underline{\textbf{89.36}} & 89.07 & 86.19 & 87.16 & \underline{88.91} & 88.88  \\

\rowcolor{gray!9}
\textit{Average} & 87.29 & 87.83 & \underline{\textbf{88.69}} & 88.48 & 80.10 & 83.17 & \underline{88.52} & 88.31 \\

\midrule
\multicolumn{9}{l}{\textbf{FinBench}} \\
\midrule                      

Banking-Short & 71.56 & 81.80 & 81.69 & \textbf{\underline{82.07}} & 78.04 & 77.33 & \underline{81.99} & 81.94 \\
Banking-Long  & 85.97 & 86.64 & 87.89 & \underline{88.40} & 88.59 & 86.41 & 88.35 & \underline{\textbf{88.89}} \\
Banking77     & 92.86 & 92.76 & 92.45 & \underline{\textbf{92.96}} & 92.04 & 90.69 & \underline{92.74} & 92.62  \\
FPB           & 84.99 & 84.99 & \underline{85.26} & 84.80 & 81.69 & 82.78 & \underline{\textbf{85.42}} & 84.60 \\
GCN           & 95.00 & \textbf{\underline{95.25}} & 95.04 & 95.08 & 94.86 & 94.91 & \underline{94.97} & 94.88 \\
Stooq         & 82.26 & 82.53 & 85.07 & \underline{\textbf{85.77}} & 81.43 & 82.04 & \underline{84.41} & 84.02 \\

\rowcolor{gray!9}
\textit{Average}       & 85.44 & 87.33 & 87.90 & \textbf{\underline{88.18}} & 86.11 & 85.69 & \underline{87.98} & 87.83 \\

\midrule
\multicolumn{9}{l}{\textbf{Other Tasks}} \\
\midrule

8TAGS      & 79.74 & 81.16 & 81.64 & \underline{\textbf{82.50}} & 75.32 & 76.43 & 81.44 & \underline{82.24} \\
BAN-PL     & 92.73 & 93.25 & 93.80 & \underline{\textbf{94.00}} & 89.39 & 90.91 & \underline{93.99} & 93.51 \\
MIPD       & 67.57 & 66.79 & 67.27 & \underline{68.28} & \underline{\textbf{70.88}} & 69.17 & 68.50 & 68.99 \\
PPC        & 87.92 & 89.78 & \underline{\textbf{89.96}} & 88.04 & 78.56 & 83.86 & \underline{89.48} & 87.20 \\
SICK-E     & 87.70 & 87.33 & \underline{88.33} & 87.88 & 82.03 & 86.58 & \underline{\textbf{88.96}} & 87.47 \\
SICK-R     & 83.47 & 84.37 & \underline{85.93} & 84.69 & 67.77 & 75.32 & \underline{\textbf{86.54}} & 84.91 \\
TwitterEMO & 70.16 & 70.51 & \underline{\textbf{70.70}} & 70.20 & 65.50 & 66.26 & \underline{70.60} & 70.35 \\
IMDB       & 91.44 & 93.55 & \underline{94.36} & 93.77 & 93.70 & 93.77 & \underline{\textbf{96.03}} & 95.93 \\
EURLEX     & 79.43 & 79.68 & 79.19 & \textbf{\underline{79.84}} & \underline{79.79} & 78.98 & 79.77 & 79.76 \\
NKJP-NER*  & 86.23 & \underline{87.79} & 84.62 & 86.84 & 86.53 & 86.67 & 87.36 & \underline{\textbf{88.66}} \\ 

\rowcolor{gray!9}
\textit{Average}     & 82.64 &  83.42 & 83.58 & \underline{83.60} & 78.95 & 80.80 & \textbf{\underline{84.27}} & 83.90
\\

\midrule
\rowcolor{gray!11}
\textbf{Short Ctx Avg}
& 84.98 & 85.95 & \underline{86.46} & \underline{86.46}
& 81.08 & 82.83 & \textbf{\underline{86.69}} & 86.43
\\

\midrule
\multicolumn{9}{l}{\textbf{LongContext}} \\
\midrule

SCOTUS-Dom          & 81.03 & 82.49 & 83.01 & \underline{83.83} & 84.20 & 82.32 & 83.99 & \underline{\textbf{85.78}} \\
SCOTUS-Dec          & 61.72 & 67.91 & \underline{72.46} & 71.39 & 76.25 & 70.97 & 68.73 & \underline{\textbf{78.21}} \\
BookSummary         & 83.43 & 84.59 & \underline{87.47} & 86.54 & 91.91 & 87.31 & \underline{\textbf{93.11}} & 91.74 \\ 
ECtHR-PL-AVA        & 58.63 & 60.23 & 58.33 & \underline{64.62} & 68.49 & 62.93 & 68.29 & \underline{\textbf{69.48}} \\
ECtHR-PL-VA         & 54.55 & 56.07 & 51.17 & \underline{61.50} & 65.97 & 63.69 & 65.27 & \underline{\textbf{67.24}} \\
\midrule
\rowcolor{gray!9}
\textbf{Long Ctx Avg}
& 67.87 & 70.26 & 70.49 & \underline{73.58}
& 77.36 & 73.44 & 75.88 & \textbf{\underline{78.49}}
\\ 

\midrule
\rowcolor{gray!11}
\textbf{Overall Average}
& 82.13 & 83.33 & 83.80 & \underline{84.31}
& 80.46 & 81.27 & 84.89 & \textbf{\underline{85.11}} \\ \\

\bottomrule
\end{tabular}
}
\vspace{-0.2cm}
\end{table*}



\subsection{LongContext Benchmark}
\label{sec:longcontext_benchmark}

We introduce \textbf{LongContext}, a five-task Polish benchmark for document
understanding beyond the standard 512-token context window. It comprises
\textsc{scotus-dom}, \textsc{scotus-dec}, \textsc{ecthr-pl-ava},
\textsc{ecthr-pl-va}, and \textsc{booksummary}. LongContext is restricted to tasks selected or constructed specifically to
evaluate long-document processing; datasets with long inputs that belong to
established benchmark groups remain in their original groups for comparability
with prior work.

The benchmark combines translated legal tasks with a specially constructed
task evaluating factual consistency between distant parts of a document. All
source datasets are based on publicly available English-language resources
and were translated into Polish using
GLM-4.6\footnote{\url{https://huggingface.co/zai-org/GLM-4.6}}. We manually spot-checked 50 randomly sampled examples from each LongContext
task. A single
annotator assessed whether the translated text was coherent and preserved the
information required for the corresponding prediction task. For BookSummary,
the inspection additionally covered claim grammaticality, unambiguity, and
consistency with the assigned label. This inspection was intended as a
quality-control check rather than a comprehensive human validation of the
benchmark.

The inspection revealed a small number of degenerate outputs containing
repetitive generation loops. We therefore applied automatic corpus-wide
filters to remove such cases and other anomalously long outputs.

\paragraph{SCOTUS.}
We derive two tasks from the publicly available SCOTUS dataset of United
States Supreme Court opinions\footnote{
\url{https://huggingface.co/datasets/pasinit/scotus}}
\citep{Spaeth2020}. \textsc{scotus-dom} is an 11-class task predicting the
legal issue area of a case, whereas \textsc{scotus-dec} is a binary task
predicting whether the ideological direction associated with the decision is
liberal or conservative. The tasks use the same court opinions but evaluate
complementary aspects of legal document understanding. Their inputs are
particularly long: 74.5\% of the test examples exceed 4K tokens when
tokenized with the Polish ModernBERT Base tokenizer.

\paragraph{ECtHR-PL.}
We construct \textsc{ecthr-pl} from the European Court of Human Rights dataset
introduced by \citet{chalkidis-et-al-2021-ecthr}.\footnote{
\url{https://huggingface.co/datasets/AUEB-NLP/ecthr_cases}}
For each case, we concatenate the sentences and paragraphs describing its
facts into a single document. We remove cases whose concatenated factual
descriptions exceed 32,000 characters to exclude extreme-length outliers and
keep the inputs within practical processing limits. The remaining documents
are translated into Polish. Using the labels provided in the original
dataset, we define two multi-label classification tasks:
\textsc{ecthr-pl-ava} predicts the articles of the European Convention on
Human Rights alleged to have been violated, whereas \textsc{ecthr-pl-va}
predicts the articles that the Court found to have been violated.

\paragraph{BookSummary.}
We construct \textsc{booksummary} from a public collection of approximately
25.7K English-language plot summaries.\footnote{
\url{https://huggingface.co/datasets/kingkangkr/book_summary_dataset}}
To obtain sufficiently long inputs, we retain the longest 25\% of the
summaries, translate them into Polish, and divide them into 70\%/10\%/20\%
training, validation, and test splits. For each summary, GLM-4.6 is prompted to generate a two- or three-sentence
claim supported by the document, with a preference for facts appearing in
the second half of the summary. The claim is then prepended to the summary. Half of the claims
are supported by the corresponding summary, while the remaining half are
modified by GLM-4.6 to be plausible but factually inconsistent. The resulting balanced binary classification task requires the model to
relate a claim at the beginning of the input to supporting or contradicting
evidence that typically appears later in the summary. The BookSummary-specific part of the manual spot-check additionally assessed
claim grammaticality, unambiguity, and consistency with the assigned label.
The claim-generation prompts are provided in
Appendix~\ref{app:booksummary_prompts}.

Most test examples in each of the three dataset families exceed the
512-token limit; full length distributions are reported in
Table~\ref{tab:dataset_length_stats}.

\subsection{Baselines}
\label{sec:baselines}

We compare Polish ModernBERT with established Polish encoder models,
including HerBERT \citep{mroczkowski-etal-2021-herbert}, Polish RoBERTa-v2
\citep{dadas2020pre}, and Polish RoBERTa-8K
\citep{dadas2026longcontext}, using Base and Large variants where available.
We also include XLM-RoBERTa \citep{conneau2019unsupervised} as an established
multilingual encoder, together with the more recent EuroBERT
\citep{boizard2025eurobertscalingmultilingualencoders} and mmBERT
\citep{marone2025mmbertmodernmultilingualencoder}, both of which provide
native long-context support. On the LongContext benchmark, 512-token models
serve as practical truncation baselines. Because the compared model families
also differ in architecture and pretraining setup, these results should not
be interpreted as a controlled ablation of context length alone.

\paragraph{Retrieval baselines.}

For retrieval evaluation, we use the Polish Information Retrieval Benchmark
(PIRB) \citep{dadas-etal-2024-pirb}, a suite of 41 Polish text-retrieval tasks.
We compare Polish ModernBERT with Polish RoBERTa-8K, EuroBERT, and mmBERT.
These encoder models are fine-tuned and evaluated using a shared
contrastive-learning protocol. We additionally include Qwen3-Embedding-0.6B
\citep{zhang2025qwen3}, BGE-M3 \citep{chen2024m3},
Jina-Embeddings-V5-Text-Small \citep{akram2026jina},
Multilingual-E5-Large \citep{yu2025arctic}, and
Snowflake-Arctic-Embed-L-v2.0 \citep{wang2024multilingual} as external
reference points. These dedicated embedding models differ in their training
data, objectives, and optimization procedures and are therefore not treated
as controlled baselines.



\subsection{Fine-Tuning and Evaluation Protocol}
\label{sec:finetuning_protocol}

For the 30 tasks in KLEJ, FinBench, Other Tasks, and LongContext, we
follow the fine-tuning protocol used for Polish RoBERTa and Polish RoBERTa-8K
\citep{dadas2020pre,dadas2026longcontext}. We fine-tune each model--task
configuration using five random seeds and report the mean test performance
according to the task's primary metric. To quantify run-to-run variability,
Table~\ref{tab:aggregate_variability} reports aggregate mean scores and sample
standard deviations across the same five runs for the principal matched
Polish-model comparisons.

The common training setup comprises 10 epochs (1 for \textsc{cbd}), a batch
size of 32, 6\% linear warm-up, and polynomial learning-rate decay. Base
models use a peak learning rate of $1\times10^{-5}$. For Large models, we
compare peak learning rates of $1\times10^{-5}$ and $3\times10^{-5}$ on the
validation set and use $3\times10^{-5}$ for the final test evaluation.
Models with standard and extended context lengths use maximum sequence
lengths of 512 and 8,192 tokens, respectively, with longer inputs truncated
to the corresponding limit. All task metrics are reported on a 0--100 scale.

\section{Results}
\label{sec:results}

\subsection{Encoder Performance}
\label{sec:encoder_results}

Tables~\ref{tab:base_results} and~\ref{tab:large_results} report results for
the Base- and Large-scale models, respectively. At the Base scale, Polish
ModernBERT achieves the strongest aggregate performance in both context
settings. The 512-token model obtains a short-context average of 84.96,
improving over Polish RoBERTa-v2 by 1.19 points, with the largest gains outside
LongContext observed on FinBench and Other Tasks. Its 8K counterpart further
increases the short-context average to 85.36 and achieves the highest overall
score of 83.99. Aggregate means and run-to-run variability for the matched Polish models are reported in Table~\ref{tab:aggregate_variability}.

The gains are also broad at the task level. Considering both context
variants, a Polish ModernBERT model attains the best Base-scale result on 17
of the 25 tasks outside LongContext, including 6 of 9 KLEJ tasks, 4 of 6
FinBench tasks, and 7 of 10 Other Tasks. The comparable short-context
averages of the 512-token and 8K variants suggest that context extension does
not degrade aggregate performance outside the dedicated LongContext
benchmark.
At the Large scale, Polish ModernBERT-512 and Polish RoBERTa-v2 obtain the
same short-context average of 86.46. Among the 512-token models, Polish
ModernBERT achieves the strongest FinBench and Other Tasks averages, while
trailing Polish RoBERTa-v2 by only 0.21 points on KLEJ. The 8K variant reaches
the highest overall score of 85.11 while retaining a short-context average
comparable to Polish RoBERTa-8K (86.43 vs.\ 86.69). Across both context
variants, the Large family provides the best result on 9 of the 25 tasks
outside LongContext, with its strongest performance concentrated in FinBench
and Other Tasks. Established Polish baselines nevertheless remain stronger
on several KLEJ tasks.

On \textsc{nkjp-ner*}, both Polish 8K model families outperform their
corresponding 512-token variants. At the Base scale, Polish RoBERTa improves
from 85.41 to 88.97, while Polish ModernBERT improves from 85.54 to 89.21.
At the Large scale, the corresponding gains are from 84.62 to 87.36 and from
86.84 to 88.66, respectively. Polish ModernBERT-8K achieves the highest score
at both scales.

Across the two scale-specific comparisons, a Polish ModernBERT variant
attains the highest task-level score in 35 of 60 cases, compared with 19 for
the Polish RoBERTa family. Polish ModernBERT also substantially outperforms
the recent multilingual EuroBERT and mmBERT variants in aggregate,
underscoring the competitiveness of Polish-specific pretraining relative to
modern multilingual encoders.

\subsection{Long-Context Performance}
\label{sec:long_context_results}

The Polish ModernBERT-8K variants attain the best score in 9 of the 10
LongContext task--scale comparisons. At the Base scale, the 8K model reaches
an average of 77.15, improving over the corresponding 512-token variant by
5.56 points and over Polish RoBERTa-8K by 9.68 points. It ranks first on all
five tasks and achieves this improvement over Polish RoBERTa-8K with 22\%
fewer parameters (149M vs.\ 190M).

At the Large scale, context extension improves the Polish ModernBERT average
from 73.58 to 78.49. The resulting model exceeds Polish RoBERTa-8K-Large by
2.61 points and ranks first on four of five tasks; the only exception is
\textsc{booksummary}, where Polish RoBERTa-8K obtains the highest score. The
largest gains over the corresponding 512-token variants are observed on
\textsc{scotus-dec}, \textsc{booksummary}, and the two \textsc{ecthr-pl}
tasks.

The gains are considerably smaller on existing datasets that contain long
examples but were not designed specifically to evaluate long-document
processing. Across \textsc{banking-long}, \textsc{mipd}, \textsc{imdb}, and
\textsc{eurlex}, the average improvement of the Polish ModernBERT-8K models
over their corresponding 512-token variants is 0.86 points at the Base scale
and 0.82 points at the Large scale, compared with 5.56 and 4.91 points on
LongContext. This contrast supports distinguishing datasets that contain long
inputs from tasks selected or constructed specifically to stress
long-document processing. To further examine how model performance changes with input length, we report
a bucket-level analysis for all LongContext tasks in
Appendix~\ref{app:length_analysis}. The corresponding figures are shown in
Figures~\ref{fig:length_bucket_scotus_books}
and~\ref{fig:length_bucket_ecthr}, while the number of examples in each bucket
is reported in Table~\ref{tab:length_bucket_counts}.

\subsection{Inference Efficiency}
\label{sec:efficiency_results}

Polish ModernBERT provides favorable quality--efficiency trade-offs. All
measurements were conducted using the same inference setup on a single
NVIDIA H100 GPU; complete results are reported in Appendix~\ref{app:efficiency}.

In the 512-token setting, Polish ModernBERT-Base improves the short-context
average over Polish RoBERTa-v2-Base from 83.77 to 84.96, while reducing
latency by 26\% and peak GPU memory usage by 54\%. At the Large scale, the
models obtain the same short-context average of 86.46, but Polish ModernBERT
reduces latency from 1.29 to 0.69 ms per sample and peak GPU memory usage
from 3,604 to 2,946 MB, corresponding to reductions of 47\% and 18\%,
respectively.

The efficiency gains persist in the 8K setting. Relative to Polish
RoBERTa-8K-Base, Polish ModernBERT-8K-Base reduces peak GPU memory usage by
24\% and latency by 6\%, while improving the LongContext average from
67.47 to 77.15. At the Large scale, Polish ModernBERT-8K reduces memory
usage by 21\% and latency by 25\%, while increasing the corresponding score
from 75.88 to 78.49.

\subsection{Retrieval Performance}


We additionally evaluate retrieval performance on PIRB, a comprehensive
Polish benchmark comprising 41 retrieval datasets. The encoder models are
fine-tuned using a shared contrastive-learning protocol, with results reported
in Table~\ref{tab:retrieval_results}. Dedicated multilingual embedding models
are included as external reference points in
Appendix~\ref{app:retrieval_embedding_references}, as they rely on different
training data, objectives, and optimization protocols. We follow the pooling, normalization, similarity, and evaluation settings of
the original PIRB protocol \citep{dadas-etal-2024-pirb}.

\begin{table}[th!]
\centering
\small
\setlength{\tabcolsep}{5pt}
\caption{Mean NDCG@10 on PIRB across 41 datasets. Bold marks the best encoder within
each size group.}
\label{tab:retrieval_results}
\begin{tabular}{lr}
\toprule
\textbf{Model} & \textbf{NDCG@10} \\
\midrule
\multicolumn{2}{l}{\textit{Smaller models ($<$300M params)}} \\
\midrule
mmBERT-small (140M)            & 48.25 \\
EuroBERT-210 (212M)           & 52.14 \\
pl-RoBERTa-8K-base (190M)   & 54.10 \\
\textbf{pl-ModernBERT-8K-base (149M)} & \textbf{55.36} \\
\midrule
\multicolumn{2}{l}{\textit{Larger models ($>$300M params)}} \\
\midrule
mmBERT-base (307M)           & 52.39 \\
EuroBERT-610M (610M)            & 56.52 \\
\textbf{pl-RoBERTa-8K-large (443M)} & \textbf{58.42} \\
pl-ModernBERT-8K-large (475M)  & 58.16 \\ 
\midrule
\end{tabular}
\vspace{-0.2cm}
\end{table}

\section{Conclusion}

We introduced \textbf{Polish ModernBERT}, a family of four Polish encoders
covering Base and Large scales and 512-token and 8K context lengths, together
with a staged pretraining procedure and a new five-task long-context
benchmark. Across 30 downstream tasks, the proposed models achieve the
strongest overall performance among the evaluated Polish encoders, with
particularly large gains on long-document understanding. Polish
ModernBERT-8K-Base outperforms the matched Polish RoBERTa-8K baseline by
9.68 points on LongContext while using 22\% fewer parameters
(149M vs.\ 190M). The models also provide favorable inference efficiency, while Polish
ModernBERT-8K-Base achieves the best PIRB result among the evaluated
encoders below 300M parameters. These results
show that modern, language-focused encoder architectures remain effective
and computationally attractive for Polish language understanding and
representation learning.

\section*{Limitations}

\paragraph{Language scope.}
Our experiments focus exclusively on Polish. Although the staged pretraining
procedure proved effective in this setting, its effectiveness may not transfer
directly to languages with different linguistic properties, data availability,
or tokenizer requirements. Evaluating the same procedure across additional
languages would be necessary to establish its broader generality.

\paragraph{Evaluation coverage.}
The evaluation suite is dominated by classification tasks, particularly
single-label classification, and includes only one sequence-labeling task.
Although this setup follows earlier evaluations of Polish encoders
\citep{dadas2026longcontext}, it does not fully represent the range of
applications for encoder models, such as reranking, span extraction, semantic
search, or structured prediction. Broader evaluation across these task types
would provide a more complete assessment of model capabilities.

\paragraph{Translated and generated evaluation data.}
Most LongContext tasks are derived from machine-translated English datasets,
while BookSummary additionally relies on LLM-generated claims. Although we
manually spot-checked 50 examples from each task and filtered degenerate
outputs, this assessment was conducted by a single annotator and was intended
as a quality-control check rather than a comprehensive human validation.
Translation artifacts, model-specific generation patterns, or occasional
label inconsistencies may therefore remain.

Moreover, four LongContext tasks are derived from publicly available English
datasets, and their source documents and labels may have appeared in the
pretraining data of some evaluated models. Translating the datasets into
Polish reduces direct lexical overlap but does not eliminate the possibility
of cross-lingual data contamination. BookSummary uses newly generated claims,
although its underlying plot summaries are also publicly available. A
larger-scale evaluation involving multiple annotators and agreement
measurements, as well as the development of originally Polish long-document
datasets, would strengthen the benchmark.

\paragraph{LongContext domain coverage.}
LongContext is weighted toward legal documents, reflecting an important
real-world setting in which long-context processing is particularly relevant.
However, this domain concentration may limit the generalizability of the
benchmark conclusions. Future extensions should include additional
long-document domains, such as finance, science, administration, and news.

\paragraph{Model-scale comparisons.}
The Base and Large variants use different tokenizers. Consequently,
differences between model scales reflect both model capacity and
tokenization and should not be interpreted as a controlled scaling study.

\section*{Ethical Considerations}

The models and benchmarks introduced in this work are intended for research
on Polish language understanding and document representation. Several
LongContext tasks are derived from publicly available legal and
human-rights datasets. Although these resources contain court documents,
the released benchmark follows the structure of the corresponding public
source datasets and is not intended for identifying individuals or making
decisions about specific cases.

Machine translation and LLM-based claim generation may introduce systematic
linguistic or factual artifacts. Moreover, performance on legal and
human-rights classification tasks should not be interpreted as evidence that
the models are suitable for autonomous legal decision-making. Model outputs
may reflect biases present in the pretraining and evaluation data and should
be independently verified in high-stakes applications.

Because the legal source documents may contain names or other identifying
details already present in the public source datasets, the released benchmark
should not be used for identifying individuals or profiling specific cases.

\section*{Acknowledgments}
This work was supported by the Gaia AI Factory project, funded by the European Union under Grant Agreement No. 101314359 through the EuroHPC Joint Undertaking (EuroHPC JU).

We gratefully acknowledge Polish high-performance computing infrastructure PLGrid (HPC Center: ACK Cyfronet AGH) for providing computer facilities and support within computational grant no. PLG/2025/018315

\bibliography{custom.bib}

@article{zhang2025qwen3,
  title={Qwen3 embedding: Advancing text embedding and reranking through foundation models},
  author={Zhang, Yanzhao and Li, Mingxin and Long, Dingkun and Zhang, Xin and Lin, Huan and Yang, Baosong and Xie, Pengjun and Yang, An and Liu, Dayiheng and Lin, Junyang and others},
  journal={arXiv preprint arXiv:2506.05176},
  year={2025}
}

@inproceedings{chen2024m3,
  title={M3-embedding: Multi-linguality, multi-functionality, multi-granularity text embeddings through self-knowledge distillation},
  author={Chen, Jianlyu and Xiao, Shitao and Zhang, Peitian and Luo, Kun and Lian, Defu and Liu, Zheng},
  booktitle={Findings of the association for computational linguistics: ACL 2024},
  pages={2318--2335},
  year={2024}
}

@inproceedings{wenzek-etal-2020-ccnet,
    title = "{CCN}et: Extracting High Quality Monolingual Datasets from Web Crawl Data",
    author = "Wenzek, Guillaume  and
      Lachaux, Marie-Anne  and
      Conneau, Alexis  and
      Chaudhary, Vishrav  and
      Guzm{\'a}n, Francisco  and
      Joulin, Armand  and
      Grave, Edouard",
    editor = "Calzolari, Nicoletta  and
      B{\'e}chet, Fr{\'e}d{\'e}ric  and
      Blache, Philippe  and
      Choukri, Khalid  and
      Cieri, Christopher  and
      Declerck, Thierry  and
      Goggi, Sara  and
      Isahara, Hitoshi  and
      Maegaard, Bente  and
      Mariani, Joseph  and
      Mazo, H{\'e}l{\`e}ne  and
      Moreno, Asuncion  and
      Odijk, Jan  and
      Piperidis, Stelios",
    booktitle = "Proceedings of the Twelfth Language Resources and Evaluation Conference",
    month = may,
    year = "2020",
    address = "Marseille, France",
    publisher = "European Language Resources Association",
    url = "https://aclanthology.org/2020.lrec-1.494/",
    pages = "4003--4012",
    language = "eng",
    ISBN = "979-10-95546-34-4"
}

@misc{Spaeth2020,
  author = {Spaeth, Harold J. and Epstein, Lee and Martin, Andrew D. and Segal, Jeffrey A. and Ruger, Theodore J. and Benesh, Sara C. and Nelson, Michael J.},
  year = {2020},
  title = {Supreme Court Database, Version 2020 Release 01},
  howpublished = {Available at \url{https://scdb.la.psu.edu/data/2020-release-01/}}
}

@inproceedings{longpre-etal-2024-pretrainers,
    title = "A Pretrainer{'}s Guide to Training Data: Measuring the Effects of Data Age, Domain Coverage, Quality, {\&} Toxicity",
    author = "Longpre, Shayne  and
      Yauney, Gregory  and
      Reif, Emily  and
      Lee, Katherine  and
      Roberts, Adam  and
      Zoph, Barret  and
      Zhou, Denny  and
      Wei, Jason  and
      Robinson, Kevin  and
      Mimno, David  and
      Ippolito, Daphne",
    editor = "Duh, Kevin  and
      Gomez, Helena  and
      Bethard, Steven",
    booktitle = "Proceedings of the 2024 Conference of the North American Chapter of the Association for Computational Linguistics: Human Language Technologies (Volume 1: Long Papers)",
    month = jun,
    year = "2024",
    address = "Mexico City, Mexico",
    publisher = "Association for Computational Linguistics",
    url = "https://aclanthology.org/2024.naacl-long.179/",
    doi = "10.18653/v1/2024.naacl-long.179",
    pages = "3245--3276"
}

@misc{olmo20252olmo2furious,
      title={2 OLMo 2 Furious}, 
      author={Team OLMo and Pete Walsh and Luca Soldaini and Dirk Groeneveld and Kyle Lo and Shane Arora and Akshita Bhagia and Yuling Gu and Shengyi Huang and Matt Jordan and Nathan Lambert and Dustin Schwenk and Oyvind Tafjord and Taira Anderson and David Atkinson and Faeze Brahman and Christopher Clark and Pradeep Dasigi and Nouha Dziri and Allyson Ettinger and Michal Guerquin and David Heineman and Hamish Ivison and Pang Wei Koh and Jiacheng Liu and Saumya Malik and William Merrill and Lester James V. Miranda and Jacob Morrison and Tyler Murray and Crystal Nam and Jake Poznanski and Valentina Pyatkin and Aman Rangapur and Michael Schmitz and Sam Skjonsberg and David Wadden and Christopher Wilhelm and Michael Wilson and Luke Zettlemoyer and Ali Farhadi and Noah A. Smith and Hannaneh Hajishirzi},
      year={2025},
      eprint={2501.00656},
      archivePrefix={arXiv},
      primaryClass={cs.CL},
      url={https://arxiv.org/abs/2501.00656}, 
}

@inproceedings{ankner-etal-2024-dynamic,
    title = "Dynamic Masking Rate Schedules for {MLM} Pretraining",
    author = "Ankner, Zachary  and
      Saphra, Naomi  and
      Blalock, Davis  and
      Frankle, Jonathan  and
      Leavitt, Matthew",
    editor = "Graham, Yvette  and
      Purver, Matthew",
    booktitle = "Proceedings of the 18th Conference of the European Chapter of the Association for Computational Linguistics (Volume 2: Short Papers)",
    month = mar,
    year = "2024",
    address = "St. Julian{'}s, Malta",
    publisher = "Association for Computational Linguistics",
    url = "https://aclanthology.org/2024.eacl-short.42/",
    doi = "10.18653/v1/2024.eacl-short.42",
    pages = "477--487"
}

@inproceedings{joulin-etal-2017-bag,
    title = "Bag of Tricks for Efficient Text Classification",
    author = "Joulin, Armand  and
      Grave, Edouard  and
      Bojanowski, Piotr  and
      Mikolov, Tomas",
    editor = "Lapata, Mirella  and
      Blunsom, Phil  and
      Koller, Alexander",
    booktitle = "Proceedings of the 15th Conference of the {E}uropean Chapter of the Association for Computational Linguistics: Volume 2, Short Papers",
    month = apr,
    year = "2017",
    address = "Valencia, Spain",
    publisher = "Association for Computational Linguistics",
    url = "https://aclanthology.org/E17-2068/",
    pages = "427--431"
}

@article{joulin2016fasttext,
  title={FastText.zip: Compressing text classification models},
  author={Joulin, Armand and Grave, Edouard and Bojanowski, Piotr and Douze, Matthijs and J{\'e}gou, H{\'e}rve and Mikolov, Tomas},
  journal={arXiv preprint arXiv:1612.03651},
  year={2016}
}

@inproceedings{heafield2011kenlm,
  title={KenLM: Faster and smaller language model queries},
  author={Heafield, Kenneth},
  booktitle={Proceedings of the sixth workshop on statistical machine translation},
  pages={187--197},
  year={2011}
}

@inbook{Ptaszynski_Pieciukiewicz_Dybala_2019, 
 title={Results of the PolEval 2019 Shared Task 6 : first dataset and Open Shared Task for automatic cyberbullying detection in Polish Twitter}, 
 ISBN={978-83-63159-28-3}, 
 url={https://ruj.uj.edu.pl/xmlui/handle/item/152265}, 
 booktitle={Proceedings of the PolEval 2019 Workshop}, 
 publisher={Polska Akademia Nauk},
 author={Ptaszynski, Michal and Pieciukiewicz, Agata and Dybała, Paweł},
 year={2019}, 
 pages={89–110}, 
 language={eng} }

@inproceedings{ogrodniczuk-kopec-2014-polish,
    title = "The {P}olish Summaries Corpus",
    author = "Ogrodniczuk, Maciej  and
      Kope{\'c}, Mateusz",
    editor = "Calzolari, Nicoletta  and
      Choukri, Khalid  and
      Declerck, Thierry  and
      Loftsson, Hrafn  and
      Maegaard, Bente  and
      Mariani, Joseph  and
      Moreno, Asuncion  and
      Odijk, Jan  and
      Piperidis, Stelios",
    booktitle = "Proceedings of the Ninth International Conference on Language Resources and Evaluation ({LREC}'14)",
    month = may,
    year = "2014",
    address = "Reykjavik, Iceland",
    publisher = "European Language Resources Association (ELRA)",
    url = "https://aclanthology.org/L14-1145/",
    pages = "3712--3715"
}

@inproceedings{kocon2019multi,
  title={Multi-level sentiment analysis of PolEmo 2.0: Extended corpus of multi-domain consumer reviews},
  author={Kocon, Jan and Mi{\l}kowski, Piotr and Za{\'s}ko-Zieli{\'n}ska, Monika},
  booktitle={Proceedings of the 23rd Conference on Computational Natural Language Learning (CoNLL)},
  pages={980--991},
  year={2019}
}

@inproceedings{marcinczuk2013open,
  title={Open dataset for development of polish question answering systems},
  author={Marcinczuk, Micha{\l} and Ptak, Marcin and Radziszewski, Adam and Piasecki, Maciej},
  booktitle={Proceedings of the 6th Language \& Technology Conference: Human Language Technologies as a Challenge for Computer Science and Linguistics, Wydawnictwo Poznanskie, Fundacja Uniwersytetu im. Adama Mickiewicza},
  year={2013}
}

@inproceedings{przepiorkowski-etal-2010-recent,
    title = "Recent Developments in the {N}ational {C}orpus of {P}olish",
    author = "Przepi{\'o}rkowski, Adam  and
      G{\'o}rski, Rafa{\l} L.  and
      {\L}azi{\'n}ski, Marek  and
      P{\k{e}}zik, Piotr",
    editor = "Calzolari, Nicoletta  and
      Choukri, Khalid  and
      Maegaard, Bente  and
      Mariani, Joseph  and
      Odijk, Jan  and
      Piperidis, Stelios  and
      Rosner, Mike  and
      Tapias, Daniel",
    booktitle = "Proceedings of the Seventh International Conference on Language Resources and Evaluation ({LREC}'10)",
    month = may,
    year = "2010",
    address = "Valletta, Malta",
    publisher = "European Language Resources Association (ELRA)",
    url = "https://aclanthology.org/L10-1097/"
}

@inproceedings{Loshchilov2017DecoupledWD,
  title={{Decoupled Weight Decay Regularization}},
  author={Ilya Loshchilov and Frank Hutter},
  booktitle={International Conference on Learning Representations},
  year={2017},
  url={https://api.semanticscholar.org/CorpusID:53592270}
}

@inproceedings{wortsman2023stable,
author = {Wortsman, Mitchell and Dettmers, Tim and Zettlemoyer, Luke and Morcos, Ari and Farhadi, Ali and Schmidt, Ludwig},
title = {Stable and low-precision training for large-scale vision-language models},
year = {2023},
publisher = {Curran Associates Inc.},
address = {Red Hook, NY, USA},
booktitle = {Proceedings of the 37th International Conference on Neural Information Processing Systems},
articleno = {451},
numpages = {28},
location = {New Orleans, LA, USA},
series = {NIPS '23}
}

@inproceedings{3495724.3497174,
author = {Zaheer, Manzil and Guruganesh, Guru and Dubey, Avinava and Ainslie, Joshua and Alberti, Chris and Ontanon, Santiago and Pham, Philip and Ravula, Anirudh and Wang, Qifan and Yang, Li and Ahmed, Amr},
title = {Big bird: transformers for longer sequences},
year = {2020},
isbn = {9781713829546},
publisher = {Curran Associates Inc.},
address = {Red Hook, NY, USA},
booktitle = {Proceedings of the 34th International Conference on Neural Information Processing Systems},
articleno = {1450},
numpages = {15},
location = {Vancouver, BC, Canada},
series = {NIPS '20}
}

@article{Beltagy2020LongformerTL,
  title={Longformer: The Long-Document Transformer},
  author={Iz Beltagy and Matthew E. Peters and Arman Cohan},
  journal={ArXiv},
  year={2020},
  volume={abs/2004.05150},
  url={https://api.semanticscholar.org/CorpusID:215737171}
}

@article{akram2026jina,
  title={jina-embeddings-v5-text: Task-targeted embedding distillation},
  author={Akram, Mohammad Kalim and Sturua, Saba and Havriushenko, Nastia and Herreros, Quentin and G{\"u}nther, Michael and Werk, Maximilian and Xiao, Han},
  journal={arXiv preprint arXiv:2602.15547},
  year={2026}
}

@inproceedings{yu2025arctic,
  title={Arctic-embed 2.0: Multilingual retrieval without compromise},
  author={Yu, Puxuan and Merrick, Luke and Nuti, Gaurav and Campos, Daniel F},
  booktitle={Second Conference on Language Modeling},
  year={2025}
}

@article{wang2024multilingual,
  title={Multilingual e5 text embeddings: A technical report},
  author={Wang, Liang and Yang, Nan and Huang, Xiaolong and Yang, Linjun and Majumder, Rangan and Wei, Furu},
  journal={arXiv preprint arXiv:2402.05672},
  year={2024}
}

@inproceedings{fan2019eli5,
  title={ELI5: Long form question answering},
  author={Fan, Angela and Jernite, Yacine and Perez, Ethan and Grangier, David and Weston, Jason and Auli, Michael},
  booktitle={Proceedings of the 57th annual meeting of the association for computational linguistics},
  pages={3558--3567},
  year={2019}
}

@misc{penedo2026finetranslations,
      title={FineTranslations}, 
      author={Guilherme Penedo and Hynek Kydl{\'\i}{\v{c}}ek and Amir Hossein Kargaran and Leandro von Werra},
      year={2026},
      publisher = {Hugging Face},
      journal = {Hugging Face repository},
      howpublished = {\url{https://huggingface.co/datasets/HuggingFaceFW/finetranslations}}
}

@inproceedings{khashabi2021gooaq,
  title={GooAQ: Open question answering with diverse answer types},
  author={Khashabi, Daniel and Ng, Amos and Khot, Tushar and Sabharwal, Ashish and Hajishirzi, Hannaneh and Callison-Burch, Chris},
  booktitle={Findings of the Association for Computational Linguistics: EMNLP 2021},
  pages={421--433},
  year={2021}
}

@inproceedings{dadas2024assessing,
  title={Assessing generalization capability of text ranking models in Polish},
  author={Dadas, S{\l}awomir and Gr\k{e}bowiec, Ma{\l}gorzata},
  booktitle={International Conference on Artificial Intelligence and Soft Computing},
  pages={37--49},
  year={2024},
  organization={Springer}
}

@inproceedings{rybak2023maupqa,
  title={MAUPQA: Massive automatically-created Polish question answering dataset},
  author={Rybak, Piotr},
  booktitle={Proceedings of the 9th Workshop on Slavic Natural Language Processing 2023 (SlavicNLP 2023)},
  pages={11--16},
  year={2023}
}

@inproceedings{wojtasik2024beir,
  title={{BEIR}-{PL}: Zero Shot Information Retrieval Benchmark for the {P}olish Language},
  author={Wojtasik, Konrad and Wo{\l}owiec, Kacper and Shishkin, Vadim and Janz, Arkadiusz and Piasecki, Maciej},
  booktitle={Proceedings of the 2024 Joint International Conference on Computational Linguistics, Language Resources and Evaluation (LREC-COLING 2024)},
  pages={2149--2160},
  year={2024}
}

@inproceedings{kobylinski2023poleval,
  title={PolEval 2022/23 challenge tasks and results},
  author={Kobyli{\'n}ski, {\L}ukasz and Ogrodniczuk, Maciej and Rybak, Piotr and Przyby{\l}a, Piotr and P{\k{e}}zik, Piotr and Miko{\l}ajczyk, Agnieszka and Janowski, Wojciech and Marci{\'n}czuk, Micha{\l} and Smywi{\'n}ski-Pohl, Aleksander},
  booktitle={2023 18th Conference on Computer Science and Intelligence Systems (FedCSIS)},
  pages={1243--1250},
  year={2023},
  organization={IEEE}
}

@inproceedings{dadas2020pre,
  title={Pre-training polish transformer-based language models at scale},
  author={Dadas, S{\l}awomir and Pere{\l}kiewicz, Micha{\l} and Po{\'s}wiata, Rafa{\l}},
  booktitle={International Conference on Artificial Intelligence and Soft Computing},
  pages={301--314},
  year={2020},
  organization={Springer}
}

@misc{dadas2026longcontext,
  title={Long-Context Encoder Models for Polish Language Understanding},
  author={Sławomir Dadas and Rafał Poświata and Marek Kozłowski and Małgorzata Grębowiec and Michał Perełkiewicz and Paweł Klimiuk and Przemysław Boruta},
  year={2026},
  eprint={2603.12191},
  archivePrefix={arXiv},
  primaryClass={cs.CL},
  url={https://arxiv.org/abs/2603.12191}
}

@misc{boizard2025eurobertscalingmultilingualencoders,
      title={EuroBERT: Scaling Multilingual Encoders for European Languages}, 
      author={Nicolas Boizard and Hippolyte Gisserot-Boukhlef and Duarte M. Alves and André Martins and Ayoub Hammal and Caio Corro and Céline Hudelot and Emmanuel Malherbe and Etienne Malaboeuf and Fanny Jourdan and Gabriel Hautreux and João Alves and Kevin El-Haddad and Manuel Faysse and Maxime Peyrard and Nuno M. Guerreiro and Patrick Fernandes and Ricardo Rei and Pierre Colombo},
      year={2025},
      eprint={2503.05500},
      archivePrefix={arXiv},
      primaryClass={cs.CL},
      url={https://arxiv.org/abs/2503.05500}, 
}

@misc{marone2025mmbertmodernmultilingualencoder,
      title={{mmBERT: A Modern Multilingual Encoder with Annealed Language Learning}}, 
      author={Marc Marone and Orion Weller and William Fleshman and Eugene Yang and Dawn Lawrie and Benjamin Van Durme},
      year={2025},
      eprint={2509.06888},
      archivePrefix={arXiv},
      primaryClass={cs.CL},
      url={https://arxiv.org/abs/2509.06888}, 
}

@inproceedings{devlin-etal-2019-bert,
    title = "{BERT}: Pre-training of Deep Bidirectional Transformers for Language Understanding",
    author = "Devlin, Jacob  and
      Chang, Ming-Wei  and
      Lee, Kenton  and
      Toutanova, Kristina",
    editor = "Burstein, Jill  and
      Doran, Christy  and
      Solorio, Thamar",
    booktitle = "Proceedings of the 2019 Conference of the North {A}merican Chapter of the Association for Computational Linguistics: Human Language Technologies, Volume 1 (Long and Short Papers)",
    month = jun,
    year = "2019",
    address = "Minneapolis, Minnesota",
    publisher = "Association for Computational Linguistics",
    url = "https://aclanthology.org/N19-1423/",
    doi = "10.18653/v1/N19-1423",
    pages = "4171--4186"
}

@misc{liu2019roberta,
  author = {Liu, Yinhan and Ott, Myle and Goyal, Naman and Du, Jingfei and Joshi, Mandar and Chen, Danqi and Levy, Omer and Lewis, Mike and Zettlemoyer, Luke and Stoyanov, Veselin},
  title = {{RoBERTa: A Robustly Optimized BERT Pretraining Approach}},
  url = {http://arxiv.org/abs/1907.11692},
  year = 2019
}

@article{DBLP:journals/corr/abs-2006-03654,
  author       = {Pengcheng He and
                  Xiaodong Liu and
                  Jianfeng Gao and
                  Weizhu Chen},
  title        = {{DeBERTa: Decoding-enhanced {BERT} with Disentangled Attention}},
  journal      = {CoRR},
  volume       = {abs/2006.03654},
  year         = {2020},
  url          = {https://arxiv.org/abs/2006.03654},
  eprinttype   = {arXiv},
  eprint       = {2006.03654},
  bibsource    = {dblp computer science bibliography, https://dblp.org}
}

@inproceedings{warner-etal-2025-smarter,
    title = "Smarter, Better, Faster, Longer: A Modern Bidirectional Encoder for Fast, Memory Efficient, and Long Context Finetuning and Inference",
    author = {Warner, Benjamin  and
      Chaffin, Antoine  and
      Clavi{\'e}, Benjamin  and
      Weller, Orion  and
      Hallstr{\"o}m, Oskar  and
      Taghadouini, Said  and
      Gallagher, Alexis  and
      Biswas, Raja  and
      Ladhak, Faisal  and
      Aarsen, Tom  and
      Adams, Griffin Thomas  and
      Howard, Jeremy  and
      Poli, Iacopo},
    editor = "Che, Wanxiang  and
      Nabende, Joyce  and
      Shutova, Ekaterina  and
      Pilehvar, Mohammad Taher",
    booktitle = "Proceedings of the 63rd Annual Meeting of the Association for Computational Linguistics (Volume 1: Long Papers)",
    month = jul,
    year = "2025",
    address = "Vienna, Austria",
    publisher = "Association for Computational Linguistics",
    url = "https://aclanthology.org/2025.acl-long.127/",
    doi = "10.18653/v1/2025.acl-long.127",
    pages = "2526--2547",
    ISBN = "979-8-89176-251-0"
}

@misc{breton2025neobertnextgenerationbert,
      title={NeoBERT: A Next-Generation BERT}, 
      author={Lola Le Breton and Quentin Fournier and Mariam El Mezouar and John X. Morris and Sarath Chandar},
      year={2025},
      eprint={2502.19587},
      archivePrefix={arXiv},
      primaryClass={cs.CL},
      url={https://arxiv.org/abs/2502.19587}, 
}

@inproceedings{NEURIPS2023_095a6917,
 author = {Portes, Jacob and Trott, Alexander and Havens, Sam and King, Daniel and Venigalla, Abhinav and Nadeem, Moin and Sardana, Nikhil and Khudia, Daya and Frankle, Jonathan},
 booktitle = {Advances in Neural Information Processing Systems},
 editor = {A. Oh and T. Naumann and A. Globerson and K. Saenko and M. Hardt and S. Levine},
 pages = {3106--3130},
 publisher = {Curran Associates, Inc.},
 title = {{MosaicBERT: A Bidirectional Encoder Optimized for Fast Pretraining}},
 url = {https://proceedings.neurips.cc/paper_files/paper/2023/file/095a6917768712b7ccc61acbeecad1d8-Paper-Conference.pdf},
 volume = {36},
 year = {2023}
}

@article{Trker2025TabiBERTAL,
  title={{TabiBERT}: A Large-Scale ModernBERT Foundation Model and Unified Benchmarking Framework for Turkish},
  author={Melik¸sah T{\"u}rker and A. Ebrar Kiziloglu and Onur G{\"u}ng{\"o}r and Susan {\"U}sk{\"u}darli},
  journal={ArXiv},
  year={2025},
  volume={abs/2512.23065},
  url={https://api.semanticscholar.org/CorpusID:284488324}
}

@inproceedings{pellicer-rinaldo-2026-norberto,
    title = "{N}or{BERT}o: A {M}odern{BERT} Model Trained for {P}ortuguese with 331 Billion Tokens Corpus",
    author = "Pellicer, Lucas F. A. O.  and
      Rinaldo, Guilherme",
    editor = "Souza, Marlo  and
      de-Dios-Flores, Iria  and
      Santos, Diana  and
      Freitas, Larissa  and
      Souza, Jackson Wilke da Cruz  and
      Ribeiro, Eug{\'e}nio",
    booktitle = "Proceedings of the 17th International Conference on Computational Processing of {P}ortuguese ({PROPOR} 2026) - Vol. 1",
    month = apr,
    year = "2026",
    address = "Salvador, Brazil",
    publisher = "Association for Computational Linguistics",
    url = "https://aclanthology.org/2026.propor-1.18/",
    pages = "183--193",
    ISBN = "979-8-89176-387-6"
}

@inproceedings{haltiuk-smywinski-pohl-2025-path,
    title = "On the Path to Make {U}krainian a High-Resource Language",
    author = "Haltiuk, Mykola  and
      Smywi{\'n}ski-Pohl, Aleksander",
    editor = "Romanyshyn, Mariana",
    booktitle = "Proceedings of the Fourth Ukrainian Natural Language Processing Workshop (UNLP 2025)",
    month = jul,
    year = "2025",
    address = "Vienna, Austria (online)",
    publisher = "Association for Computational Linguistics",
    url = "https://aclanthology.org/2025.unlp-1.14/",
    doi = "10.18653/v1/2025.unlp-1.14",
    pages = "120--130",
    ISBN = "979-8-89176-269-5"
}

@misc{shmidman2025neodictabertpushingfrontierbert,
      title={{NeoDictaBERT: Pushing the Frontier of BERT models for Hebrew}}, 
      author={Shaltiel Shmidman and Avi Shmidman and Moshe Koppel},
      year={2025},
      eprint={2510.20386},
      archivePrefix={arXiv},
      primaryClass={cs.CL},
      url={https://arxiv.org/abs/2510.20386}, 
}

@misc{reunamo2025pretrainingfinnishmodernberts,
      title={{Pretraining Finnish ModernBERTs}}, 
      author={Akseli Reunamo and Laura-Maria Peltonen and Hans Moen and Sampo Pyysalo},
      year={2025},
      eprint={2511.09213},
      archivePrefix={arXiv},
      primaryClass={cs.CL},
      url={https://arxiv.org/abs/2511.09213}, 
}

@inproceedings{clark2020electra,
  title = {{ELECTRA}: Pre-training Text Encoders as Discriminators Rather Than Generators},
  author = {Kevin Clark and Minh-Thang Luong and Quoc V. Le and Christopher D. Manning},
  booktitle = {ICLR},
  year = {2020},
  url = {https://openreview.net/pdf?id=r1xMH1BtvB}
}

@inproceedings{rybak-etal-2020-klej,
    title = "{KLEJ}: Comprehensive Benchmark for {P}olish Language Understanding",
    author = "Rybak, Piotr  and
      Mroczkowski, Robert  and
      Tracz, Janusz  and
      Gawlik, Ireneusz",
    editor = "Jurafsky, Dan  and
      Chai, Joyce  and
      Schluter, Natalie  and
      Tetreault, Joel",
    booktitle = "Proceedings of the 58th Annual Meeting of the Association for Computational Linguistics",
    month = jul,
    year = "2020",
    address = "Online",
    publisher = "Association for Computational Linguistics",
    url = "https://aclanthology.org/2020.acl-main.111/",
    doi = "10.18653/v1/2020.acl-main.111",
    pages = "1191--1201"
}

@inproceedings{mroczkowski-etal-2021-herbert,
    title = "{H}er{BERT}: Efficiently Pretrained Transformer-based Language Model for {P}olish",
    author = "Mroczkowski, Robert  and
      Rybak, Piotr  and
      Wr{\'o}blewska, Alina  and
      Gawlik, Ireneusz",
    editor = "Babych, Bogdan  and
      Kanishcheva, Olga  and
      Nakov, Preslav  and
      Piskorski, Jakub  and
      Pivovarova, Lidia  and
      Starko, Vasyl  and
      Steinberger, Josef  and
      Yangarber, Roman  and
      Marci{\'n}czuk, Micha{\l}  and
      Pollak, Senja  and
      P{\v{r}}ib{\'a}{\v{n}}, Pavel  and
      Robnik-{\v{S}}ikonja, Marko",
    booktitle = "Proceedings of the 8th Workshop on Balto-Slavic Natural Language Processing",
    month = apr,
    year = "2021",
    address = "Kyiv, Ukraine",
    publisher = "Association for Computational Linguistics",
    url = "https://aclanthology.org/2021.bsnlp-1.1/",
    pages = "1--10"
}

@inproceedings{wroblewska2017polish,
  title={Polish evaluation dataset for compositional distributional semantics models},
  author={Wr{\'o}blewska, Alina and Krasnowska-Kiera{\'s}, Katarzyna},
  booktitle={Proceedings of the 55th Annual Meeting of the Association for Computational Linguistics (Volume 1: Long Papers)},
  pages={784--792},
  year={2017}
}

@INPROCEEDINGS{7078567,
  author={Rennie, Steven J. and Goel, Vaibhava and Thomas, Samuel},
  booktitle={2014 IEEE Spoken Language Technology Workshop (SLT)}, 
  title={Annealed dropout training of deep networks}, 
  year={2014},
  volume={},
  number={},
  pages={159-164},
  doi={10.1109/SLT.2014.7078567}}

@article{10.1109/TASLP.2021.3124365,
author = {Cui, Yiming and Che, Wanxiang and Liu, Ting and Qin, Bing and Yang, Ziqing},
title = {Pre-Training With Whole Word Masking for Chinese BERT},
year = {2021},
issue_date = {2021},
publisher = {IEEE Press},
volume = {29},
issn = {2329-9290},
url = {https://doi.org/10.1109/TASLP.2021.3124365},
doi = {10.1109/TASLP.2021.3124365},
journal = {IEEE/ACM Trans. Audio, Speech and Lang. Proc.},
month = nov,
pages = {3504–3514},
numpages = {11}
}

@inproceedings{wettig-etal-2023-mask,
    title = "Should You Mask 15{\%} in Masked Language Modeling?",
    author = "Wettig, Alexander  and
      Gao, Tianyu  and
      Zhong, Zexuan  and
      Chen, Danqi",
    editor = "Vlachos, Andreas  and
      Augenstein, Isabelle",
    booktitle = "Proceedings of the 17th Conference of the European Chapter of the Association for Computational Linguistics",
    month = may,
    year = "2023",
    address = "Dubrovnik, Croatia",
    publisher = "Association for Computational Linguistics",
    url = "https://aclanthology.org/2023.eacl-main.217/",
    doi = "10.18653/v1/2023.eacl-main.217",
    pages = "2985--3000"
}

@inproceedings{dadas-etal-2024-pirb,
    title = "{PIRB}: A Comprehensive Benchmark of {P}olish Dense and Hybrid Text Retrieval Methods",
    author = "Dadas, Slawomir  and
      Pere{\l}kiewicz, Micha{\l}  and
      Po{\'s}wiata, Rafa{\l}",
    editor = "Calzolari, Nicoletta  and
      Kan, Min-Yen  and
      Hoste, Veronique  and
      Lenci, Alessandro  and
      Sakti, Sakriani  and
      Xue, Nianwen",
    booktitle = "Proceedings of the 2024 Joint International Conference on Computational Linguistics, Language Resources and Evaluation (LREC-COLING 2024)",
    month = may,
    year = "2024",
    address = "Torino, Italia",
    publisher = "ELRA and ICCL",
    url = "https://aclanthology.org/2024.lrec-main.1117/",
    pages = "12761--12774"
}

@inproceedings{pires-etal-2019-multilingual,
    title = "How Multilingual is Multilingual {BERT}?",
    author = "Pires, Telmo  and  Schlinger, Eva  and  Garrette, Dan",
    booktitle = "Proceedings of the 57th Annual Meeting of the Association for Computational Linguistics",
    year = "2019",
    pages = "4996--5001",
}

@article{conneau2019unsupervised,
  title={Unsupervised Cross-lingual Representation Learning at Scale},
  author={Conneau, Alexis and Khandelwal, Kartikay and Goyal, Naman and Chaudhary, Vishrav and Wenzek, Guillaume and Guzm{\'a}n, Francisco and Grave, Edouard and Ott, Myle and Zettlemoyer, Luke and Stoyanov, Veselin},
  journal={arXiv preprint arXiv:1911.02116},
  year={2019}
}

@inproceedings{ruder-etal-2021-xtreme,
    title = "{XTREME}-{R}: Towards More Challenging and Nuanced Multilingual Evaluation",
    author = "Ruder, Sebastian  and
      Constant, Noah  and
      Botha, Jan  and
      Siddhant, Aditya  and
      Firat, Orhan  and
      Fu, Jinlan  and
      Liu, Pengfei  and
      Hu, Junjie  and
      Garrette, Dan  and
      Neubig, Graham  and
      Johnson, Melvin",
    editor = "Moens, Marie-Francine  and
      Huang, Xuanjing  and
      Specia, Lucia  and
      Yih, Scott Wen-tau",
    booktitle = "Proceedings of the 2021 Conference on Empirical Methods in Natural Language Processing",
    month = nov,
    year = "2021",
    address = "Online and Punta Cana, Dominican Republic",
    publisher = "Association for Computational Linguistics",
    url = "https://aclanthology.org/2021.emnlp-main.802/",
    doi = "10.18653/v1/2021.emnlp-main.802",
    pages = "10215--10245"
}

@misc{hu2020xtrememassivelymultilingualmultitask,
      title={{XTREME: A Massively Multilingual Multi-task Benchmark for Evaluating Cross-lingual Generalization}}, 
      author={Junjie Hu and Sebastian Ruder and Aditya Siddhant and Graham Neubig and Orhan Firat and Melvin Johnson},
      year={2020},
      eprint={2003.11080},
      archivePrefix={arXiv},
      primaryClass={cs.CL},
      url={https://arxiv.org/abs/2003.11080}, 
}

@inproceedings{kudo-richardson-2018-sentencepiece,
    title = "{S}entence{P}iece: A simple and language independent subword tokenizer and detokenizer for Neural Text Processing",
    author = "Kudo, Taku  and
      Richardson, John",
    editor = "Blanco, Eduardo  and
      Lu, Wei",
    booktitle = "Proceedings of the 2018 Conference on Empirical Methods in Natural Language Processing: System Demonstrations",
    month = nov,
    year = "2018",
    address = "Brussels, Belgium",
    publisher = "Association for Computational Linguistics",
    url = "https://aclanthology.org/D18-2012/",
    doi = "10.18653/v1/D18-2012",
    pages = "66--71"
}

@article{Loshchilov2016SGDRSG,
  title={{SGDR: Stochastic Gradient Descent with Restarts}},
  author={Ilya Loshchilov and Frank Hutter},
  journal={ArXiv},
  year={2016},
  volume={abs/1608.03983},
  url={https://api.semanticscholar.org/CorpusID:15884797}
}

@InProceedings{chalkidis-et-al-2021-ecthr,
    title = "Paragraph-level Rationale Extraction through Regularization: A case study on European Court of Human Rights Cases",
    author = "Chalkidis, Ilias and Fergadiotis, Manos and Tsarapatsanis, Dimitrios and Aletras, Nikolaos and Androutsopoulos, Ion and Malakasiotis, Prodromos",
    booktitle = "Proceedings of the Annual Conference of the North American Chapter of the Association for Computational Linguistics",
    year = "2021",
    address = "Mexico City, Mexico",
    publisher = "Association for Computational Linguistics"
}

@inproceedings{casanueva-etal-2020-efficient,
    title = "Efficient Intent Detection with Dual Sentence Encoders",
    author = "Casanueva, I{\~n}igo  and
      Tem{\v{c}}inas, Tadas  and
      Gerz, Daniela  and
      Henderson, Matthew  and
      Vuli{\'c}, Ivan",
    editor = "Wen, Tsung-Hsien  and
      Celikyilmaz, Asli  and
      Yu, Zhou  and
      Papangelis, Alexandros  and
      Eric, Mihail  and
      Kumar, Anuj  and
      Casanueva, I{\~n}igo  and
      Shah, Rushin",
    booktitle = "Proceedings of the 2nd Workshop on Natural Language Processing for Conversational AI",
    month = jul,
    year = "2020",
    address = "Online",
    publisher = "Association for Computational Linguistics",
    url = "https://aclanthology.org/2020.nlp4convai-1.5/",
    doi = "10.18653/v1/2020.nlp4convai-1.5",
    pages = "38--45",
}

@article{malo2014good,
  title={Good debt or bad debt: Detecting semantic orientations in economic texts},
  author={Malo, Pekka and Sinha, Ankur and Korhonen, Pekka and Wallenius, Jyrki and Takala, Pyry},
  journal={Journal of the Association for Information Science and Technology},
  volume={65},
  number={4},
  pages={782--796},
  year={2014},
  publisher={Wiley Online Library}
}

@inproceedings{sinha2021impact,
  title={Impact of news on the commodity market: Dataset and results},
  author={Sinha, Ankur and Khandait, Tanmay},
  booktitle={Future of Information and Communication Conference},
  pages={589--601},
  year={2021},
  organization={Springer}
}

@inproceedings{dadas-etal-2020-evaluation,
    title = "Evaluation of Sentence Representations in {P}olish",
    author = "Dadas, Slawomir  and
      Pere{\l}kiewicz, Micha{\l}  and
      Po{\'s}wiata, Rafa{\l}",
    editor = "Calzolari, Nicoletta  and
      B{\'e}chet, Fr{\'e}d{\'e}ric  and
      Blache, Philippe  and
      Choukri, Khalid  and
      Cieri, Christopher  and
      Declerck, Thierry  and
      Goggi, Sara  and
      Isahara, Hitoshi  and
      Maegaard, Bente  and
      Mariani, Joseph  and
      Mazo, H{\'e}l{\`e}ne  and
      Moreno, Asuncion  and
      Odijk, Jan  and
      Piperidis, Stelios",
    booktitle = "Proceedings of the Twelfth Language Resources and Evaluation Conference",
    month = may,
    year = "2020",
    address = "Marseille, France",
    publisher = "European Language Resources Association",
    url = "https://aclanthology.org/2020.lrec-1.207/",
    pages = "1674--1680",
    language = "eng",
    ISBN = "979-10-95546-34-4"
}

@inproceedings{kolos-etal-2024-ban,
    title = "{BAN}-{PL}: A {P}olish Dataset of Banned Harmful and Offensive Content from Wykop.pl Web Service",
    author = "Kolos, Anna  and
      Okulska, Inez  and
      G{\l}{\k{a}}bi{\'n}ska, Kinga  and
      Karlinska, Agnieszka  and
      Wisnios, Emilia  and
      Ellerik, Pawe{\l}  and
      Pra{\l}at, Andrzej",
    editor = "Calzolari, Nicoletta  and
      Kan, Min-Yen  and
      Hoste, Veronique  and
      Lenci, Alessandro  and
      Sakti, Sakriani  and
      Xue, Nianwen",
    booktitle = "Proceedings of the 2024 Joint International Conference on Computational Linguistics, Language Resources and Evaluation (LREC-COLING 2024)",
    month = may,
    year = "2024",
    address = "Torino, Italia",
    publisher = "ELRA and ICCL",
    url = "https://aclanthology.org/2024.lrec-main.190/",
    pages = "2107--2118"
}

@inproceedings{modzelewski-etal-2024-mipd,
    title = "{MIPD}: Exploring Manipulation and Intention In a Novel Corpus of {P}olish Disinformation",
    author = "Modzelewski, Arkadiusz  and
      Da San Martino, Giovanni  and
      Savov, Pavel  and
      Wilczy{\'n}ska, Magdalena Anna  and
      Wierzbicki, Adam",
    editor = "Al-Onaizan, Yaser  and
      Bansal, Mohit  and
      Chen, Yun-Nung",
    booktitle = "Proceedings of the 2024 Conference on Empirical Methods in Natural Language Processing",
    month = nov,
    year = "2024",
    address = "Miami, Florida, USA",
    publisher = "Association for Computational Linguistics",
    url = "https://aclanthology.org/2024.emnlp-main.1103/",
    doi = "10.18653/v1/2024.emnlp-main.1103",
    pages = "19769--19785"
}

@inproceedings{dadas2022training,
  title={Training effective neural sentence encoders from automatically mined paraphrases},
  author={Dadas, S{\l}awomir},
  booktitle={2022 IEEE International Conference on Systems, Man, and Cybernetics (SMC)},
  pages={371--378},
  year={2022},
  organization={IEEE}
}

@inproceedings{bogdanowicz2023twitteremo,
  title={Twitteremo: Annotating emotions and sentiment in polish twitter},
  author={Bogdanowicz, Stanis{\l}aw and Cwynar, Hanna and Zwierzchowska, Aleksandra and Klamra, Cezary and Kiera{\'s}, Witold and Kobyli{\'n}ski, {\L}ukasz},
  booktitle={International Conference on Computational Science},
  pages={212--220},
  year={2023},
  organization={Springer}
}

@inproceedings{maas2011learning,
  title={Learning word vectors for sentiment analysis},
  author={Maas, Andrew and Daly, Raymond E and Pham, Peter T and Huang, Dan and Ng, Andrew Y and Potts, Christopher},
  booktitle={Proceedings of the 49th annual meeting of the association for computational linguistics: Human language technologies},
  pages={142--150},
  year={2011}
}

@inproceedings{chalkidis-etal-2019-large,
    title = "Large-Scale Multi-Label Text Classification on {EU} Legislation",
    author = "Chalkidis, Ilias  and Fergadiotis, Manos  and Malakasiotis, Prodromos  and Androutsopoulos, Ion",
    booktitle = "Proceedings of the 57th Annual Meeting of the Association for Computational Linguistics",
    year = "2019",
    address = "Florence, Italy",
    publisher = "Association for Computational Linguistics",
    url = "https://www.aclweb.org/anthology/P19-1636",
    doi = "10.18653/v1/P19-1636",
    pages = "6314--6322"
}

\appendix


\section{Corpus Composition and Cleaning}
\label{app:corpus_composition}

\subsection{Domain Composition}
\label{app:domain_composition}

To characterize the diversity of the pretraining data, we estimate the domain
composition of the main corpus after cleaning and deduplication. Documents are
assigned to topical categories using a lightweight multinomial Naive Bayes
classifier operating on lemmatized document tokens represented as
TF--IDF-weighted bag-of-words features. The classifier was trained on
approximately 15 GB of Polish text assembled from public sources that could be
mapped to specific domains and achieved 78\% validation accuracy.

Because several fine-grained labels are closely related and the classifier
provides only approximate domain assignments, we aggregate them into broader
categories to obtain more robust and interpretable corpus-level estimates.
For reporting, predicted document-level labels are aggregated into the eight
broad categories shown in Table~\ref{tab:domain_composition}, and category
shares are computed by token count. The original classifier predicts 19
fine-grained domains, which are aggregated as follows: technology and social
networks into \textit{Technology / Internet}; science and engineering and
biomedicine into \textit{Science / Biomedicine}; humanities and social
sciences, history, religion, and art into \textit{Humanities}; finance and
e-commerce into \textit{Finance / Commerce}; lifestyle and entertainment,
food, sport, and automotive content into \textit{Lifestyle}; and housing and
construction, agriculture, and other content into \textit{Home / Other}.
News remains a separate category, while law is reported under
\textit{Law / Public Affairs}.

The resulting statistics should therefore be interpreted as approximate
corpus-level estimates rather than manual document-level annotations. No
single domain dominates the token-weighted corpus. Common Crawl contains a
larger share of news and technology-related content, while the curated corpus
contains higher proportions of legal, scientific, and humanities text;
FineTranslations is more evenly distributed across news, technology, science,
and humanities.

\begin{table}[th]
\centering
\small
\setlength{\tabcolsep}{4pt}
\caption{
Approximate domain composition of the main pretraining corpus after cleaning
and deduplication. Values denote token percentages within each corpus
component; the Total column is computed as a token-weighted average using the
component sizes reported in Table~\ref{tab:pretraining_data}. Curated denotes
the curated Polish corpus, CC denotes Common Crawl, and FineTran denotes
FineTranslations.
}
\label{tab:domain_composition}
\begin{tabular}{lrrrr}
\toprule
\textbf{Domain} &
\textbf{Curated} &
\textbf{CC} &
\textbf{FineTran} &
\textbf{Total} \\
\midrule
News                    & 8.0  & 28.0 & 20.0 & 18.7 \\
Technology / Internet   & 10.0 & 25.0 & 18.0 & 17.7 \\
Science / Biomedicine   & 20.0 & 10.0 & 18.0 & 15.8 \\
Law / Public Affairs    & 22.0 & 6.0  & 8.0  & 12.2 \\
Humanities              & 18.0 & 8.0  & 14.0 & 13.2 \\
Finance / Commerce      & 7.0  & 10.0 & 8.0  & 8.4 \\
Lifestyle               & 8.0  & 9.0  & 6.0  & 7.8 \\
Home / Other            & 7.0  & 4.0  & 8.0  & 6.2 \\
\bottomrule
\end{tabular}
\end{table}


\subsection{Corpus Cleaning}
\label{app:corpus_cleaning}

All corpus components are processed with a shared cleaning and filtering
pipeline. Documents are first segmented into sentences using NLTK-based
sentence segmentation extended with Polish abbreviation lists. We then apply
punctuation and whitespace normalization based on CCNet-style rules \citep{wenzek-etal-2020-ccnet} and remove
URLs. Sentences longer than 100 characters with less than 40\% alphabetic
characters are discarded, and documents shorter than 500 characters are
removed.

Language filtering is performed at the sentence level with a FastText-based
language-identification model \cite{joulin-etal-2017-bag,joulin2016fasttext}. Sentences not assigned to Polish with sufficient
confidence are removed, while the threshold is kept low to avoid discarding
valid Polish sentences that may be difficult to classify reliably.

We additionally apply classifier-based quality filtering. The quality filter is
a lightweight binary random-forest classifier implemented with scikit-learn. It
uses 23 numeric features describing character-, word-, and sentence-level
statistics of each document, including the proportion of letters, digits,
whitespace and punctuation, capitalization patterns, word-length statistics,
sentence-length statistics, and repetition features. The classifier was trained
on 2,520 manually labeled Polish documents, consisting of 1,417 high-quality
and 1,103 low-quality examples, using an 80/20 train--validation split. It
achieved 96\% validation accuracy. Documents classified as low quality are
removed from the corpus.

Finally, we apply KenLM-based perplexity filtering \citep{heafield2011kenlm}. Perplexity is computed with
a lightweight statistical language model, and documents exceeding the selected
perplexity threshold are discarded. This step is intended to remove documents
with abnormal language-model scores, including noisy extraction artifacts,
poorly encoded texts, and other low-quality content that may not be captured
by the preceding filters.

\paragraph{Deduplication}

After cleaning, we apply exact and near-duplicate removal. Exact duplicates are
removed using SHA-256 document hashes stored in a Bloom filter, which provides
a memory-efficient way to track previously observed documents at corpus scale.

Near-duplicate removal is based on MinHash locality-sensitive hashing. Each
document is converted into a set of unique word trigrams, represented with a
128-permutation MinHash signature. Candidate near-duplicate pairs are retrieved
with LSH and grouped when their estimated Jaccard similarity exceeds 0.7.
Within each near-duplicate cluster, we retain the document with the highest
available quality score. If no quality score is available, we retain the
earliest processed document. The final deduplicated corpus is then used to
construct the main pretraining corpus and the stage-specific mixtures described
in Section~\ref{sec:pretraining_corpus}.

\section{Recipe-Selection Experiments}
\label{app:recipe_selection}

\subsection{Direct Transfer of the ModernBERT Recipe}
\label{app:org_recipe}

As an initial baseline, we trained a Polish ModernBERT-Base-8K model by
directly adapting the original ModernBERT pretraining recipe
\citep{warner-etal-2025-smarter}. The baseline followed the original sequence
of main pretraining, long-context continuation, and final learning-rate
annealing as closely as possible, while using the same cleaned Polish data
sources as the final models.

This experiment was intended to test whether the original recipe could be
transferred to Polish without additional recipe selection. We therefore did
not apply the staged modifications introduced in our final schedule, including
the four-stage 512-token curriculum, stage-specific corpus refinement,
progressive masking-ratio reduction, and independently restarted learning-rate
schedules.

\begin{table}[th!]
\centering
\small
\caption{
Validation performance of the direct-transfer ModernBERT recipe, averaged over
five fine-tuning runs.
}
\label{tab:org_recipe}
\begin{tabular}{lc}
\toprule
\textbf{Task group} & \textbf{Average} \\
\midrule
KLEJ (9 tasks)             & 83.76 \\
FinBench (6 tasks)         & 83.69 \\
KLEJ + FinBench (15 tasks) & 83.73 \\
\bottomrule
\end{tabular}
\end{table}

As shown in Table~\ref{tab:org_recipe}, the direct-transfer baseline achieved
83.76 on KLEJ, 83.69 on FinBench, and 83.73 on the combined 15-task
development suite. While this represents a strong initial baseline, the staged
search identified configurations with higher validation performance,
motivating the final recipe-selection procedure described in
Appendix~\ref{app:pretraining_recipe_selection}.

\subsection{Pretraining Recipe Selection}
\label{app:pretraining_recipe_selection}

\paragraph{Common training settings.}
Unless stated otherwise, all stages use decoupled StableAdamW
\citep{wortsman2023stable,Loshchilov2017DecoupledWD} with
$\beta_1=0.9$, $\beta_2=0.98$, $\epsilon=10^{-6}$, and weight decay
$10^{-6}$. Bias and normalization parameters are excluded from weight decay.
Training uses BF16 mixed precision and dynamic masking. At each stage
transition, only the selected model parameters are carried over; the optimizer
and learning-rate scheduler states are reinitialized. The data loader is also
reinitialized with a new random seed, yielding a new document order, while
masks are sampled dynamically throughout training. Gradient clipping with a
threshold of 1.0 is applied in all stages except Stage~I. Embedding and MLP
dropout are disabled throughout training. Unless stated otherwise, the selected
configuration at each stage uses an independently initialized cosine schedule
with 6\% linear warm-up.

\paragraph{Checkpoint selection.}
Intermediate checkpoints are evaluated every 10\% of the stage budget,
starting at 30\%. For each checkpoint and task, validation scores are averaged
over five fine-tuning runs; the resulting task-level scores are then
macro-averaged across KLEJ. For each configuration, we report the highest KLEJ
average observed across its evaluated checkpoints. The best-performing
configuration and checkpoint within each stage are propagated to the subsequent
stage. Table~\ref{tab:recipe_selection} summarizes the explored configurations
and selected variants.

\begin{table*}[t]
\centering
\small
\setlength{\tabcolsep}{4pt}
\caption{
Recipe-selection experiments conducted on the Base model. Each stage starts
from the checkpoint selected in the preceding stage. For each configuration,
we report the highest KLEJ validation average observed across the evaluated
checkpoints; task-level scores are averaged over five fine-tuning runs before
macro-averaging across KLEJ. Bold indicates the configuration selected for
continuation. For WSD variants, the percentage denotes the fraction of the
stage allocated to learning-rate decay.
}
\label{tab:recipe_selection}
\begin{tabular}{llllllll}
\toprule
\textbf{Stage} & \textbf{Variant} & \textbf{Corpus} & \textbf{Objective}
& \textbf{Masking} & \textbf{Peak LR} & \textbf{Attn. dropout}
& \textbf{Avg. KLEJ} \\
\midrule
I   & WSD-5\%              & Full      & MLM & 0.30 & $8\times10^{-4}$ & 0.10 & 82.87 \\
I   & WSD-10\%             & Full      & MLM & 0.30 & $8\times10^{-4}$ & 0.10 & 83.05 \\
I   & WSD-15\%             & Full      & MLM & 0.30 & $8\times10^{-4}$ & 0.10 & 83.18 \\
I   & \textbf{Cosine}      & Full      & MLM & 0.30 & $8\times10^{-4}$ & 0.10 & \textbf{83.42} \\
\midrule
II  & MLM-30               & Full      & MLM & 0.30 & $2\times10^{-4}$ & 0.10 & 83.88 \\
II  & MLM-15               & Full      & MLM & 0.15 & $2\times10^{-4}$ & 0.10 & 84.16 \\
II  & \textbf{WWM-25}      & Full      & WWM & 0.25 & $2\times10^{-4}$ & 0.10 & \textbf{84.48} \\
II  & WWM-15               & Full      & WWM & 0.15 & $2\times10^{-4}$ & 0.10 & 84.29 \\
\midrule
III & Full-corpus continuation
                              & Full      & WWM & 0.25 & $1\times10^{-4}$ & 0.10 & 84.72 \\
III & Curated corpus        & Curated   & WWM & 0.25 & $1\times10^{-4}$ & 0.10 & 85.39 \\
III & \textbf{Curated + lower masking}
                              & Curated   & WWM & 0.15 & $1\times10^{-4}$ & 0.10 & \textbf{85.62} \\
\midrule
IV  & Annealing             & Annealing & WWM & 0.08 & $5\times10^{-5}$ & 0.10 & 86.62 \\
IV  & \textbf{Annealing without dropout}
                              & Annealing & WWM & 0.08 & $5\times10^{-5}$ & 0.00 & \textbf{87.00} \\
\bottomrule
\end{tabular}
\end{table*}

\paragraph{Stage I: optimization schedule.}
We retained the original ModernBERT Base token-level MLM objective, masking
probability of 0.30, and peak learning rate of $8\times10^{-4}$
\citep{warner-etal-2025-smarter}. We compared warmup--stable--decay schedules
with decay phases covering the final 5\%, 10\%, and 15\% of training against
cosine decay with 6\% linear warm-up. All remaining settings were fixed. The
corresponding average KLEJ scores were 82.87, 83.05, 83.18, and 83.42,
respectively. Cosine decay performed best and was retained for the subsequent
stages.

Recipe selection was conducted on Base for 200K steps, with a nominal token
budget of 64.4B. The selected schedule was transferred to Large using the
original ModernBERT Large peak learning rate of $5\times10^{-4}$ and 250K
steps, corresponding to a nominal token budget of 80.5B. The final Stage~I
checkpoints initialized Stage~II.

\paragraph{Stage II: masking objective and ratio.}
Stage~II compared token-level MLM with masking probabilities of 0.30 and 0.15
against whole-word masking (WWM) with probabilities of 0.25 and 0.15
\citep{10.1109/TASLP.2021.3124365}. The corresponding average KLEJ scores were 83.88,
84.16, 84.48, and 84.29. WWM with a masking probability of 0.25 performed best
and was selected for Stage~III. This comparison was motivated by prior
evidence that the optimal masking rate can depend on the objective and
training stage \citep{wettig-etal-2023-mask,ankner-etal-2024-dynamic}.

All Stage~II variants used the full pretraining corpus, maximum sequence
length of 512, attention dropout of 0.1, and batch token capacity of
approximately 1.6M. We restarted the learning-rate schedule with 6\% linear
warm-up and cosine decay, reducing the common peak learning rate to
$2\times10^{-4}$ for both Base and Large. The stage ran for 200K steps for
Base and 250K steps for Large, corresponding to nominal token budgets of 320B
and 400B tokens. The selected checkpoints occurred at 80\% and 90\% of the
respective stage budgets.

\paragraph{Stage III: corpus and masking refinement.}
Stage~III evaluated the transition from the full corpus to the curated corpus
and a lower WWM probability. Continuing on the full corpus with WWM probability
0.25 yielded an average KLEJ score of 84.72. Replacing it with the curated
corpus increased the score to 85.39, while additionally reducing the masking
probability to 0.15 produced 85.62. The latter configuration was selected for
Stage~IV. These experiments were motivated by evidence that corpus quality and
composition can substantially affect pretraining outcomes
\citep{longpre-etal-2024-pretrainers}.

The stage used a maximum sequence length of 512, attention dropout of 0.1,
a batch token capacity of approximately 966K, and a peak learning rate of
$1\times10^{-4}$. Training continued for 100K steps, corresponding to a
nominal budget of 96.6B tokens. The selected checkpoints occurred at 70\% of
the stage budget for Base and 90\% for Large.

\paragraph{Stage IV: annealing.}
The final 512-token stage introduced the annealing corpus, reduced the WWM
probability from 0.15 to 0.08, and lowered the peak learning rate from
$1\times10^{-4}$ to $5\times10^{-5}$. This late-stage refinement follows
related pretraining strategies used in ModernBERT and OLMo~2
\citep{warner-etal-2025-smarter,olmo20252olmo2furious}. The complete annealing configuration obtained an average KLEJ score of 86.62,
compared with 85.62 for the selected Stage~III checkpoint. Because the corpus,
masking probability, and learning rate were modified jointly, this comparison
measures the effect of the complete annealing configuration rather than
isolating individual factors.

We subsequently compared attention dropout values of 0.1 and 0.0
\citep{7078567}, while keeping the corpus, objective, masking ratio, learning
rate, and training budget fixed. The corresponding average KLEJ scores were
86.62 and 87.00, respectively. Zero
attention dropout was selected. The stage used a batch token capacity of
approximately 966K and ran for 100K steps, corresponding to a nominal budget
of 96.6B tokens. The selected checkpoint occurred at the end of training for
Base and at 90\% of the stage budget for Large. These checkpoints constitute
the final 512-token models.

\paragraph{Context extension.}
The selected Stage~IV checkpoints initialized the 8K variants. We retained the
tokenizer, architecture, and pretrained parameters, increased the maximum
sequence length from 512 to 8,192, and changed the global RoPE base from
10,000 to 160,000; the local RoPE base remained 10,000. Context extension used
the dedicated long-context corpus, WWM with a masking probability of 0.30, and
zero dropout. The schedule was restarted with a peak learning rate of
$1\times10^{-4}$, 6\% linear warm-up, and cosine decay. The stage used a batch
token capacity of approximately 491K and ran for 200K steps, corresponding to
a nominal budget of 98.2B tokens. The selected checkpoints occurred at 70\%
and 80\% of the stage budget for Base and Large, respectively.









\section{Evaluation Details}
\label{app:evaluation_details}

This appendix provides additional details on the evaluation datasets, task
formulations, dataset splits, and the analysis of model performance across
input-length ranges.

\subsection{Task Overview}
\label{app:dataset_statistics}

The evaluation suite comprises 30 tasks drawn from four benchmark groups:
KLEJ, FinBench, Other Tasks, and LongContext. These tasks cover diverse
formulations and domains, including classification, regression, sequence
labeling, semantic similarity, finance, social media, legal documents, and
literary texts. Tables~\ref{tab:results_by_task_type} and
\ref{tab:results_by_domain} report macro-averaged performance grouped by task
type and domain, respectively. Table~\ref{tab:evaluation_tasks} summarizes the
complete task set together with the task type, domain, and primary evaluation
metric.

Table~\ref{tab:dataset_length_stats} reports the number of test examples,
token-length statistics, long-input shares, and unknown-token rates for all
evaluation datasets. These statistics support the distinction between
standard short-context tasks, datasets containing some long inputs, and the
LongContext tasks designed to evaluate document-level processing.

\begin{table*}[t]
\setlength{\tabcolsep}{4pt}
\centering
\small
\caption{
Overview of evaluation tasks grouped by benchmark source. Type and domain labels are used for aggregate analyses; metrics denote the primary score reported for each task.
}
\label{tab:evaluation_tasks}
\begin{tabular}{llll}
\toprule
\textbf{Task} & \textbf{Type} & \textbf{Domain} & \textbf{Metric} \\
\midrule

\multicolumn{4}{l}{\textbf{KLEJ Benchmark} \citep{rybak-etal-2020-klej}} \\
\midrule
NKJP-NER \citep{przepiorkowski-etal-2010-recent}
& single-label & mixed & Accuracy \\

CDSC-E \citep{wroblewska2017polish}
& single-label & semantics & Accuracy \\

CDSC-R \citep{wroblewska2017polish}
& regression & semantics & Spearman \\

CBD \citep{Ptaszynski_Pieciukiewicz_Dybala_2019}
& single-label & social media & Binary F1 \\

POLEMO-IN \citep{kocon2019multi}
& single-label & reviews & Accuracy \\

POLEMO-OUT \citep{kocon2019multi}
& single-label & reviews & Accuracy \\

DYK \citep{marcinczuk2013open}
& single-label & mixed & Binary F1 \\

PSC \citep{ogrodniczuk-kopec-2014-polish}
& single-label & news & Binary F1 \\

AR \citep{rybak-etal-2020-klej}
& regression & reviews & $1-\mathrm{wMAE}$ \\

\midrule
\multicolumn{4}{l}{\textbf{FinBench} \citep{dadas2026longcontext}} \\
\midrule
Banking-Short \citep{dadas2026longcontext} & single-label & finance & Accuracy \\
Banking-Long \citep{dadas2026longcontext}  & single-label & finance & Accuracy \\
Banking77 \citep{casanueva-etal-2020-efficient} & single-label & finance & Accuracy \\
FPB \citep{malo2014good} & single-label & finance & Accuracy \\
GCN \citep{sinha2021impact} & multi-label  & finance & Weighted F1 \\
Stooq \citep{dadas2026longcontext} & single-label & finance & Accuracy \\

\midrule
\multicolumn{4}{l}{\textbf{Other Tasks} \citep{dadas2026longcontext}} \\
\midrule
8TAGS \citep{dadas-etal-2020-evaluation} & single-label & soc. media & Accuracy \\
BAN-PL \citep{kolos-etal-2024-ban} & single-label & soc. media & Accuracy \\
MIPD \citep{modzelewski-etal-2024-mipd} & multi-label  & news         & Weighted F1 \\
PPC \citep{dadas2022training} & single-label & semantics    & Accuracy \\
SICK-E \citep{dadas-etal-2020-evaluation} & single-label & semantics    & Accuracy \\
SICK-R \citep{dadas-etal-2020-evaluation} & regression   & semantics    & Spearman \\
TwitterEMO \citep{bogdanowicz2023twitteremo} & multi-label  & soc. media & Weighted F1 \\
IMDB \citep{maas2011learning} & single-label & reviews      & Accuracy \\
EURLEX \citep{chalkidis-etal-2019-large} & multi-label  & law          & Weighted F1 \\
NKJP-NER* \citep{przepiorkowski-etal-2010-recent}  & seq label         & mixed      & entity-lvl micro-F1 \\

\midrule
\multicolumn{4}{l}{\textbf{LongContext (our)}} \\
\midrule
SCOTUS-Dom \citep{Spaeth2020} & single-label      & law        & Accuracy \\
SCOTUS-Dec \citep{Spaeth2020} & single-label      & law        & Accuracy \\
BookSummary     & single-label      & literature & Accuracy \\
ECtHR-PL-AVA \citep{chalkidis-et-al-2021-ecthr} & multi-label       & law        & Weighted F1 \\
ECtHR-PL-VA \citep{chalkidis-et-al-2021-ecthr} & multi-label       & law        & Weighted F1 \\

\bottomrule
\end{tabular}
\end{table*}

\begin{table*}[!th]
\centering
\small
\caption{
Token-length statistics for the evaluation datasets, computed on the test
splits using the Polish RoBERTa-v2 Base tokenizer used to train the Polish
ModernBERT Base models. Test size denotes the number of test examples. Input
length denotes the number of tokens per test example. Long-input share denotes
the percentage of test examples exceeding each token-length threshold, and UNK
denotes the percentage of tokenized pieces mapped to the unknown token.
Slash-separated task names indicate tasks sharing the same input texts.
}
\label{tab:dataset_length_stats}
\resizebox{\textwidth}{!}{
\begin{tabular}{l r rrrr rrrr r}
\toprule
\textbf{Dataset}
& \textbf{Test size}
& \multicolumn{4}{c}{\textbf{Input length (tokens)}}
& \multicolumn{4}{c}{\textbf{Long-input share (\%)}}
& \textbf{UNK (\%)} \\
\cmidrule(lr){3-6}
\cmidrule(lr){7-10}
&
& Mean
& Median
& P95
& Max
& $>512$
& $>1$K
& $>2$K
& $>4$K
& \\
\midrule

\multicolumn{11}{l}{\textbf{KLEJ}} \\
\midrule
NKJP-NER
& 2018 & 23 & 20 & 48 & 1003
& 0.1 & 0.0 & 0.0 & 0.0 & 1.2 \\

POLEMO2.0-IN
& 722 & 183 & 163 & 381 & 719
& 1.5 & 0.0 & 0.0 & 0.0 & 1.3 \\

POLEMO2.0-OUT
& 483 & 152 & 143 & 279 & 1894
& 0.4 & 0.2 & 0.0 & 0.0 & 1.5 \\

CBD
& 2000 & 31 & 30 & 54 & 80
& 0.0 & 0.0 & 0.0 & 0.0 & 10.7 \\

CDSC-E / CDSC-R
& 1000 & 36 & 35 & 54 & 90
& 0.0 & 0.0 & 0.0 & 0.0 & 0.1 \\

DYK
& 1029 & 68 & 57 & 142 & 450
& 0.0 & 0.0 & 0.0 & 0.0 & 2.2 \\

AR
& 1006 & 107 & 95 & 196 & 474
& 0.0 & 0.0 & 0.0 & 0.0 & 1.1 \\

PSC
& 1078 & 212 & 189 & 376 & 543
& 0.7 & 0.0 & 0.0 & 0.0 & 0.9 \\

\midrule
\multicolumn{11}{l}{\textbf{FinBench}} \\
\midrule
Banking-Short
& 2800 & 16 & 16 & 26 & 50
& 0.0 & 0.0 & 0.0 & 0.0 & 2.0 \\

Banking-Long
& 2800 & 548 & 377 & 1593 & 7014
& 36.7 & 12.5 & 2.4 & 0.3 & 1.9 \\

Banking77
& 3080 & 21 & 18 & 46 & 132
& 0.0 & 0.0 & 0.0 & 0.0 & 2.1 \\

FPB
& 970 & 47 & 43 & 85 & 141
& 0.0 & 0.0 & 0.0 & 0.0 & 2.1 \\

GCN
& 2264 & 19 & 18 & 31 & 51
& 0.0 & 0.0 & 0.0 & 0.0 & 0.8 \\

Stooq
& 363 & 95 & 79 & 194 & 611
& 0.3 & 0.0 & 0.0 & 0.0 & 0.8 \\

\midrule
\multicolumn{11}{l}{\textbf{Other tasks}} \\
\midrule
8TAGS
& 4372 & 20 & 17 & 38 & 100
& 0.0 & 0.0 & 0.0 & 0.0 & 1.8 \\

BAN-PL
& 4000 & 47 & 27 & 137 & 3036
& 0.3 & 0.0 & 0.0 & 0.0 & 1.9 \\

EURLEX
& 5000 & 2258 & 687 & 9508 & 243863
& 65.1 & 34.3 & 18.3 & 10.1 & 2.2 \\

IMDB
& 2000 & 1244 & 1145 & 1896 & 4509
& 100.0 & 67.5 & 1.2 & 0.2 & 2.9 \\

MIPD
& 1521 & 1156 & 790 & 3027 & 17098
& 77.0 & 37.7 & 10.8 & 2.8 & 2.3 \\

PPC
& 1000 & 21 & 18 & 40 & 99
& 0.0 & 0.0 & 0.0 & 0.0 & 0.6 \\

SICK-E / SICK-R
& 4906 & 20 & 18 & 34 & 66
& 0.0 & 0.0 & 0.0 & 0.0 & 0.1 \\

TwitterEMO
& 4000 & 40 & 34 & 80 & 224
& 0.0 & 0.0 & 0.0 & 0.0 & 7.3 \\

NKJP-NER-SENT & 36079 & 15 & 12 & 36 & 214 & 0.0 & 0.0 & 0.0 & 0.0 & 0.0 \\

NKJP-NER-DOC & 1828 & 295 & 278 & 340 & 2676 & 1.1 & 1.1 & 0.3 & 0.0 & 0.0 \\
\midrule
\multicolumn{11}{l}{\textbf{LongContext}} \\
\midrule
SCOTUS-Dom / Dec
& 931 & 5534 & 6572 & 7758 & 57573
& 95.4 & 91.8 & 86.9 & 74.5 & 2.9 \\

BookSummary
& 1248 & 1057 & 871 & 2423 & 5916
& 87.9 & 38.7 & 8.1 & 0.3 & 1.9 \\

ECtHR-PL-AVA / VA
& 819 & 1454 & 1219 & 3172 & 6363
& 83.4 & 58.6 & 23.3 & 2.4 & 1.8 \\

\bottomrule
\end{tabular}
}
\end{table*}

\begin{table}[th]
\centering
\small
\setlength{\tabcolsep}{5pt}
\caption{
Split sizes for LongContext datasets and NKJP-NER* variants. A dash indicates
that no separate validation split is used.
}
\label{tab:longcontext_splits}
\begin{tabular}{lrrrr}
\toprule
\textbf{Task} & \textbf{Train} & \textbf{Valid} & \textbf{Test} & \textbf{Total} \\
\midrule
SCOTUS-Dom       & 7,413  & 912 & 931    & 9,256 \\
SCOTUS-Dec       & 7,413  & 912 & 931    & 9,256 \\
BookSummary      & 4,460  & 620 & 1,248  & 6,328 \\
ECtHR-PL-AVA     & 7,367  & 834 & 819    & 9,020 \\
ECtHR-PL-VA      & 7,367  & 834 & 819    & 9,020 \\
\midrule
NKJP-NER-SENT    & 85,628 & --  & 36,079 & 121,707 \\
NKJP-NER-DOC     & 18,473 & --  & 1,828  & 20,301 \\
\bottomrule
\end{tabular}
\end{table}

\begin{table*}[t]
\centering
\small
\setlength{\tabcolsep}{4pt}
\caption{
Macro-average performance grouped by task type.
Averages are computed over 20 single-label, 6 multi-label,
3 regression, and 1 sequence-labeling tasks.
Best results within each model-scale block are shown in bold;
underlined values indicate the best result among models using the same context length.
}
\label{tab:results_by_task_type}

\resizebox{\textwidth}{!}{
\begin{tabular}{
l
S S S S
!{\color{gray!60}\vrule width 0.3pt}
S S S S
}
\toprule
&
\multicolumn{4}{c}{\textbf{512 tokens}} &
\multicolumn{4}{c}{\textbf{8K tokens}} \\
\cmidrule(lr){2-5}
\cmidrule(lr){6-9}

\textbf{Task type}
& {\textbf{XLM-RoBERTa(m)}}
& {\textbf{HerBERT}}
& {\textbf{pl-RoBERTa-v2}}
& {\textbf{pl-ModernBERT}}
& {\textbf{EuroBERT(m)}}
& {\textbf{mmBERT(m)}}
& {\textbf{pl-RoBERTa-8K}}
& {\textbf{pl-ModernBERT}} \\
\midrule

\multicolumn{9}{l}{\textit{Base models}} \\
\midrule

single-label 
& 81.01
& 82.63
& 84.16
& \underline{84.98}
& 77.85
& 79.72
& 84.52
& \underline{\textbf{85.95}} \\

multi-label 
& 54.15
& 63.22
& 58.14
& \underline{71.84}
& 67.45
& 70.07
& 66.34
& \underline{\textbf{74.21}} \\

regression 
& 84.74
& 86.47
& 88.35
& \underline{88.60}
& 75.09
& 80.48
& 88.70
& \underline{\textbf{88.76}} \\

sequence labeling 
& 84.36
& 85.49
& 85.41
& \underline{85.54}
& 84.76
& 83.82
& 88.97
& \underline{\textbf{89.21}} \\

\midrule
\multicolumn{9}{l}{\textit{Large models}} \\
\midrule

single-label 
& 84.28
& 85.75
& \underline{86.85}
& 86.70
& 81.84
& 82.98
& 87.05
& \underline{\textbf{87.28}} \\

multi-label 
& 70.89
& 71.42
& 70.28
& \underline{73.25}
& 74.25
& 72.29
& 71.98
& \underline{\textbf{75.12}} \\

regression 
& 88.92
& 89.54
& \underline{90.18}
& 89.63
& 81.70
& 85.20
& \underline{\textbf{90.26}}
& 89.42 \\

sequence labeling 
& 86.23
& \underline{87.79}
& 84.62
& 86.84
& 86.53
& 86.67
& 87.36
& \underline{\textbf{88.66}} \\

\bottomrule
\end{tabular}
}
\end{table*}

\begin{table*}[t]
\centering
\small
\setlength{\tabcolsep}{4pt}
\caption{
Macro-average performance grouped by domain.
Averages are computed over 3 mixed-domain, 5 semantic, 4 social-media,
4 review, 6 financial, 2 news, 5 legal, and 1 literary task.
Best available results within each model-scale block are shown in bold;
underlined values indicate the best available result among models using the same context length.
}
\label{tab:results_by_domain}

\resizebox{\textwidth}{!}{
\begin{tabular}{
l
S S S S
!{\color{gray!60}\vrule width 0.3pt}
S S S S
}
\toprule
&
\multicolumn{4}{c}{\textbf{512 tokens}} &
\multicolumn{4}{c}{\textbf{8K tokens}} \\
\cmidrule(lr){2-5}
\cmidrule(lr){6-9}

\textbf{Domain}
& {\textbf{XLM-RoBERTa(m)}}
& {\textbf{HerBERT}}
& {\textbf{pl-RoBERTa-v2}}
& {\textbf{pl-ModernBERT}}
& {\textbf{EuroBERT(m)}}
& {\textbf{mmBERT(m)}}
& {\textbf{pl-RoBERTa-8K}}
& {\textbf{pl-ModernBERT}} \\
\midrule

\multicolumn{9}{l}{\textit{Base models}} \\
\midrule

mixed 
& 80.58
& 82.81
& \underline{83.37}
& 82.27
& 68.45
& 74.36
& \underline{\textbf{84.16}}
& 83.66 \\

semantics 
& 85.28
& 87.04
& \underline{\textbf{89.01}}
& 88.53
& 75.01
& 82.93
& \underline{88.99}
& 88.92 \\

social media 
& 73.22
& 75.82
& 76.81
& \underline{77.84}
& 68.51
& 69.51
& 77.48
& \underline{\textbf{78.71}} \\

reviews 
& 86.31
& 86.60
& 87.24
& \underline{89.31}
& 84.45
& 85.73
& 88.90
& \underline{\textbf{89.51}} \\

finance 
& 83.13
& 83.65
& 85.14
& \underline{\textbf{86.95}}
& 82.38
& 83.07
& 85.98
& \underline{86.69} \\

news 
& 76.1
& 78.3
& 78.73
& \underline{82.45}
& 80.13
& 82.03
& 81.65
& \underline{\textbf{82.86}} \\

law 
& 49.57
& 60.78
& 55.56
& \underline{70.71}
& 68.56
& 67.29
& 62.50
& \underline{\textbf{75.03}} \\

literature 
& 78.53
& 81.12
& \underline{85.02}
& 83.71
& 82.27
& 82.58
& 88.96
& \underline{\textbf{90.22}} \\

\midrule
\multicolumn{9}{l}{\textit{Large models}} \\
\midrule

mixed 
& 84.69
& 85.72
& \underline{85.08}
& 85.06
& 72.91
& 81.08
& 85.68
& \underline{\textbf{85.89}} \\

semantics 
& 89.65
& 90.25
& \underline{90.73}
& 90.07
& 81.80
& 86.52
& \underline{\textbf{90.92}}
& 89.75 \\

social media 
& 77.38
& 78.78
& 79.81
& \underline{\textbf{79.83}}
& 71.20
& 71.54
& \underline{79.81}
& 79.39 \\

reviews 
& 88.55
& 88.96
& \underline{90.27}
& 90.16
& 87.30
& 87.38
& 90.41
& \underline{\textbf{90.66}} \\

finance 
& 85.44
& 87.33
& 87.90
& \underline{\textbf{88.18}}
& 86.11
& 85.69
& \underline{87.98}
& 87.82 \\

news 
& 83.24
& 82.82
& 82.82
& \underline{83.55}
& \underline{\textbf{84.45}}
& 83.00
& 83.53
& 83.73 \\

law 
& 67.07
& 69.28
& 68.83
& \underline{72.24}
& 74.94
& 71.34
& 70.11
& \underline{\textbf{76.09}} \\

literature 
& 83.43
& 84.59
& \underline{87.47}
& 86.54
& 91.91
& 87.31
& \underline{\textbf{93.11}}
& 91.74 \\

\bottomrule
\end{tabular}
}
\end{table*}

\subsection{LongContext and NKJP-NER* Dataset Splits}
\label{app:longcontext_splits}

Table~\ref{tab:longcontext_splits} summarizes the dataset splits used for the
LongContext benchmark and the document-level NKJP-NER* analysis. For SCOTUS
and ECtHR-PL, we retain the original train, validation, and test splits after
translation and filtering. For BookSummary, we retain the longest 25\% of
summaries and split the resulting set into approximately 70\%/10\%/20\%
training, validation, and test partitions. NKJP-NER-SENT and NKJP-NER-DOC are
derived from the same underlying NKJP texts and annotations, but differ in
sentence- and document-level segmentation; no separate validation split is
used for these variants.

\subsection{Performance by Input Length}
\label{app:length_analysis}

To complement the aggregate LongContext results, we analyze performance as a
function of input length. For each task, test examples are grouped into four
length buckets, and the mean task score is computed separately within each
bucket. Input lengths are measured using the tokenizer of
\texttt{polish-roberta-base-v2} for all models, ensuring that each example is
assigned to the same bucket across model comparisons.

\begin{figure*}[t]
\centering

\begin{subfigure}[t]{0.49\textwidth}
    \centering
    \includegraphics[width=\linewidth]{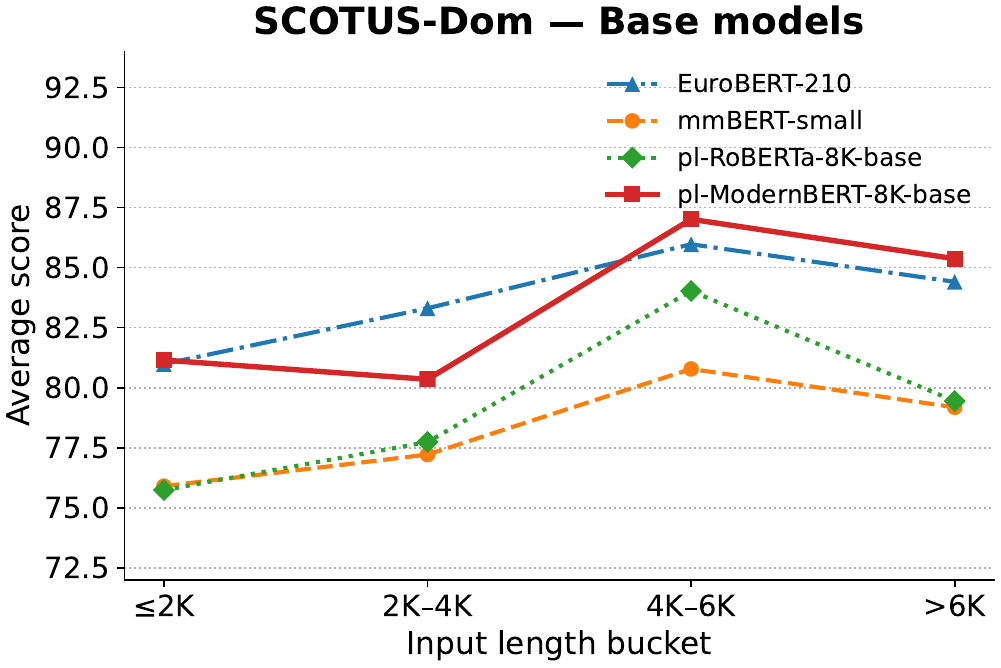}
    \caption{\textsc{scotus-dom}: Base models.}
    \label{fig:length_scotus_dom_base}
\end{subfigure}
\hfill
\begin{subfigure}[t]{0.49\textwidth}
    \centering
    \includegraphics[width=\linewidth]{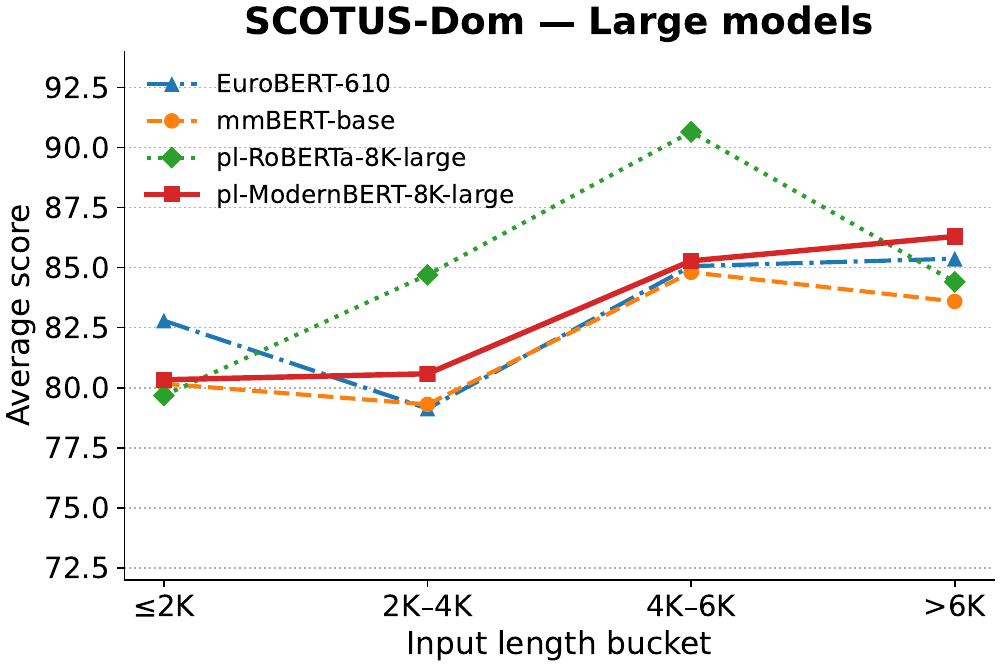}
    \caption{\textsc{scotus-dom}: Large models.}
    \label{fig:length_scotus_dom_large}
\end{subfigure}

\vspace{0.3em}

\begin{subfigure}[t]{0.49\textwidth}
    \centering
    \includegraphics[width=\linewidth]{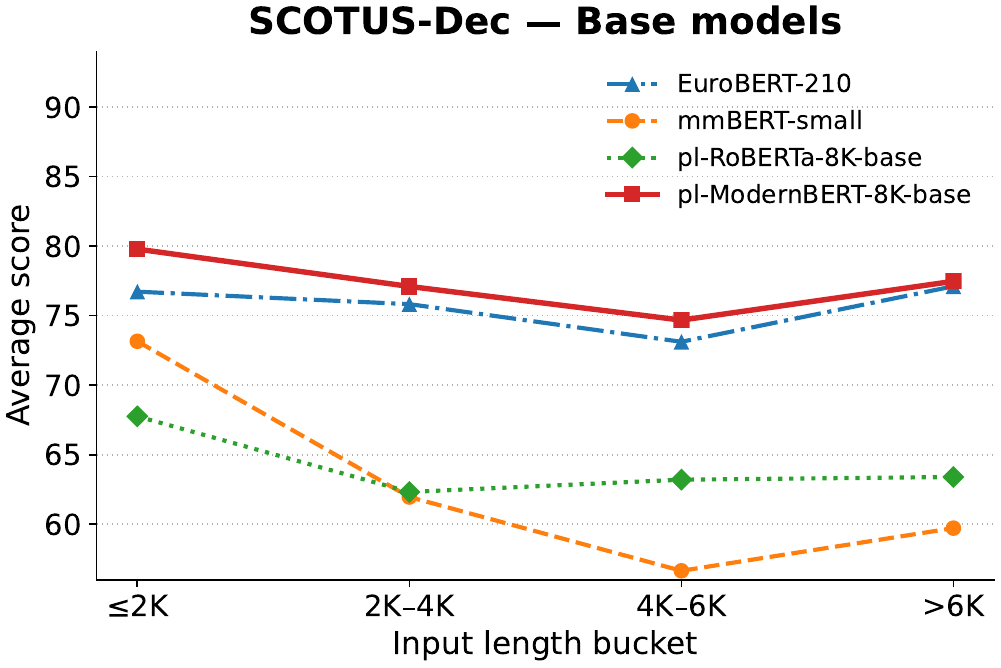}
    \caption{\textsc{scotus-dec}: Base models.}
    \label{fig:length_scotus_dec_base}
\end{subfigure}
\hfill
\begin{subfigure}[t]{0.49\textwidth}
    \centering
    \includegraphics[width=\linewidth]{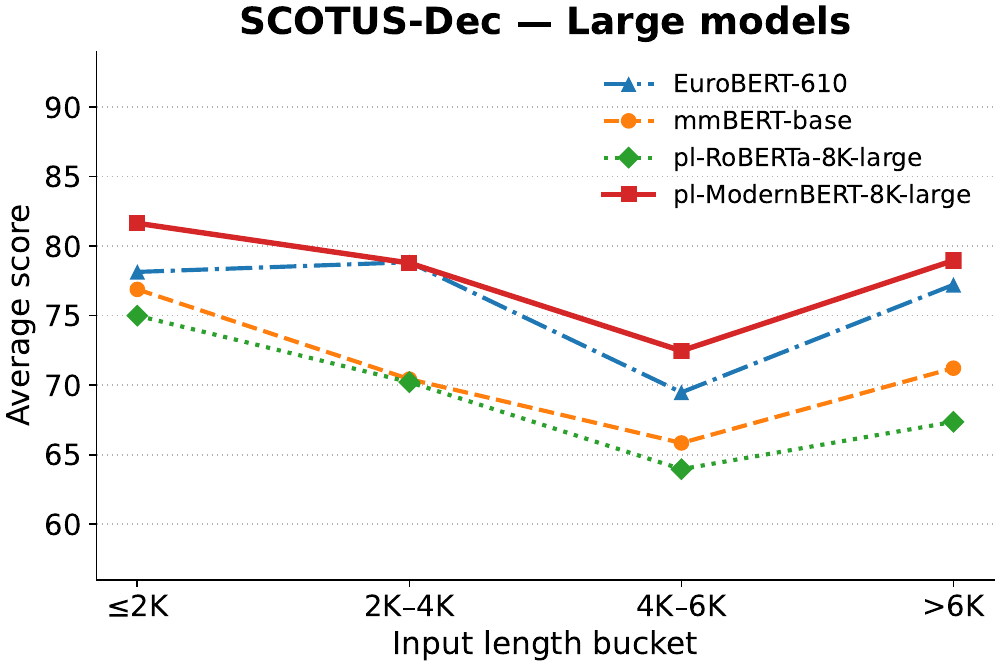}
    \caption{\textsc{scotus-dec}: Large models.}
    \label{fig:length_scotus_dec_large}
\end{subfigure}

\vspace{0.3em}

\begin{subfigure}[t]{0.49\textwidth}
    \centering
    \includegraphics[width=\linewidth]{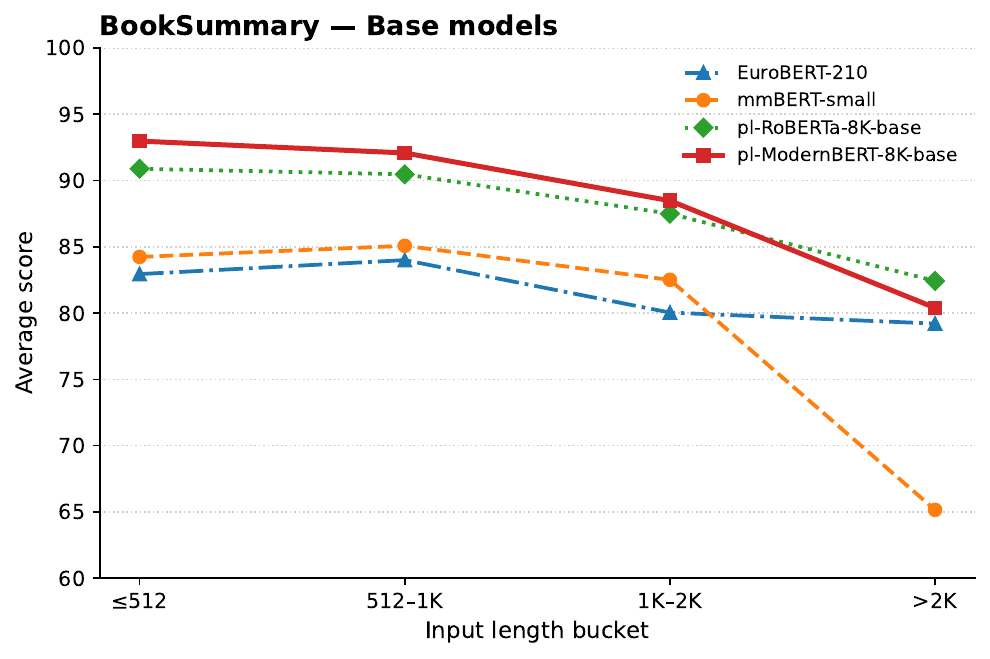}
    \caption{\textsc{booksummary}: Base models.}
    \label{fig:length_booksummary_base}
\end{subfigure}
\hfill
\begin{subfigure}[t]{0.49\textwidth}
    \centering
    \includegraphics[width=\linewidth]{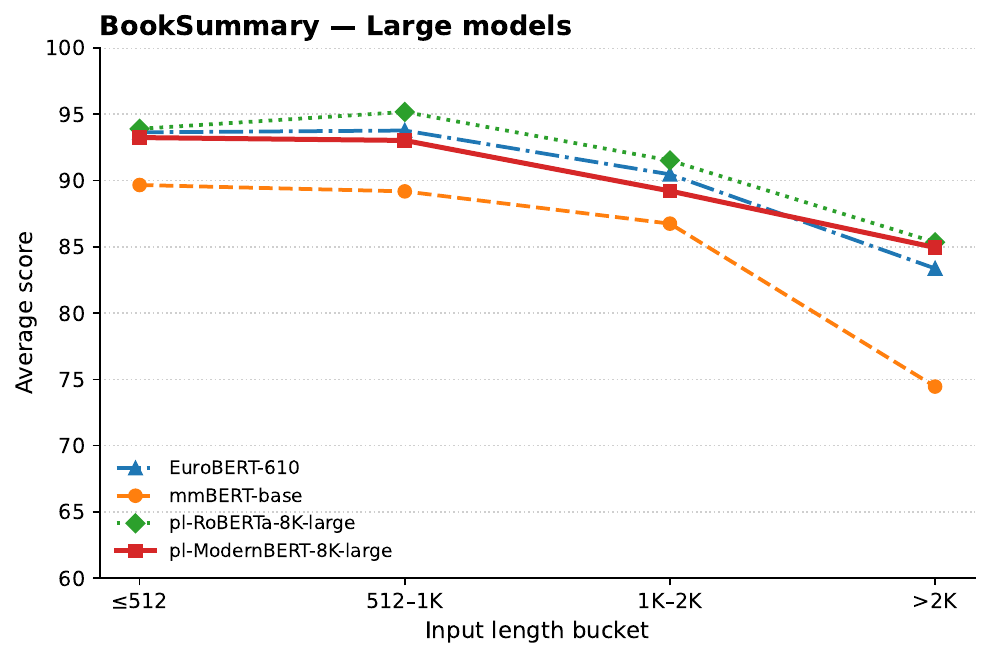}
    \caption{\textsc{booksummary}: Large models.}
    \label{fig:length_booksummary_large}
\end{subfigure}

\caption{
Performance by input length on the SCOTUS-Dom, SCOTUS-Dec, and BookSummary
LongContext tasks. Each point shows the mean test score for examples within
the corresponding length bucket. Results are presented separately for Base
and Large models. Bucket definitions and example counts are provided in
Table~\ref{tab:length_bucket_counts}.
}
\label{fig:length_bucket_scotus_books}
\end{figure*}

\begin{figure*}[t]
\centering

\begin{subfigure}[t]{0.49\textwidth}
    \centering
    \includegraphics[width=\linewidth]{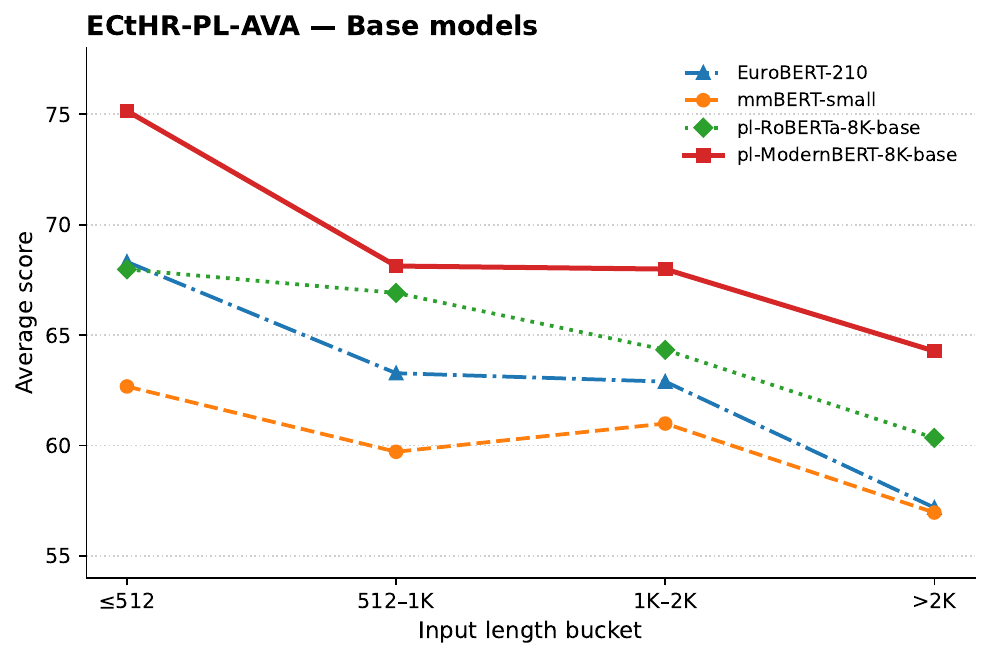}
    \caption{\textsc{ecthr-pl-ava}: Base models.}
    \label{fig:length_ecthr_ava_base}
\end{subfigure}
\hfill
\begin{subfigure}[t]{0.49\textwidth}
    \centering
    \includegraphics[width=\linewidth]{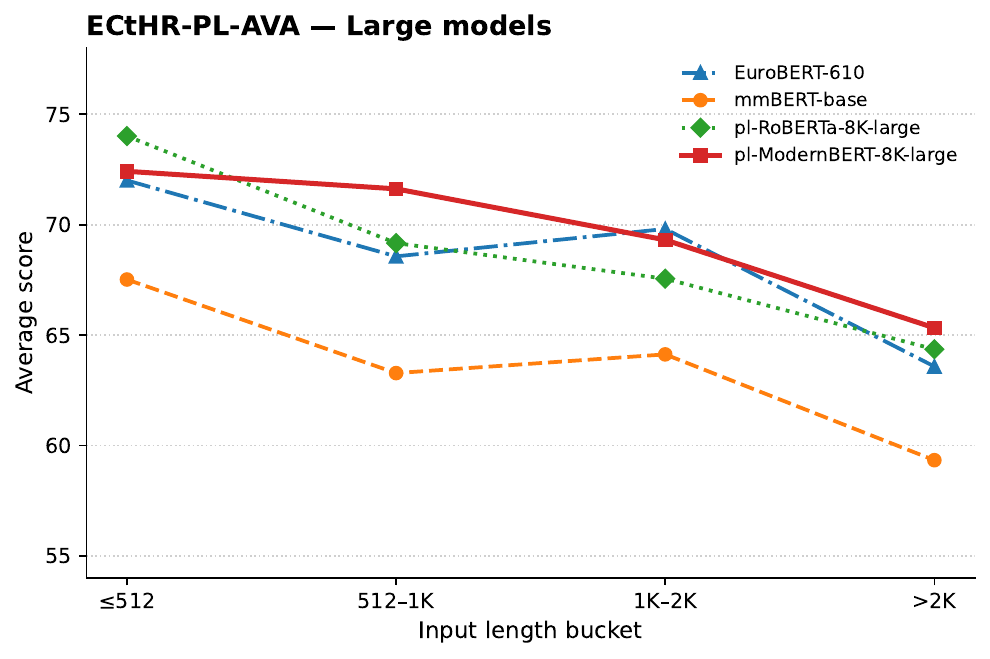}
    \caption{\textsc{ecthr-pl-ava}: Large models.}
    \label{fig:length_ecthr_ava_large}
\end{subfigure}

\vspace{0.3em}

\begin{subfigure}[t]{0.49\textwidth}
    \centering
    \includegraphics[width=\linewidth]{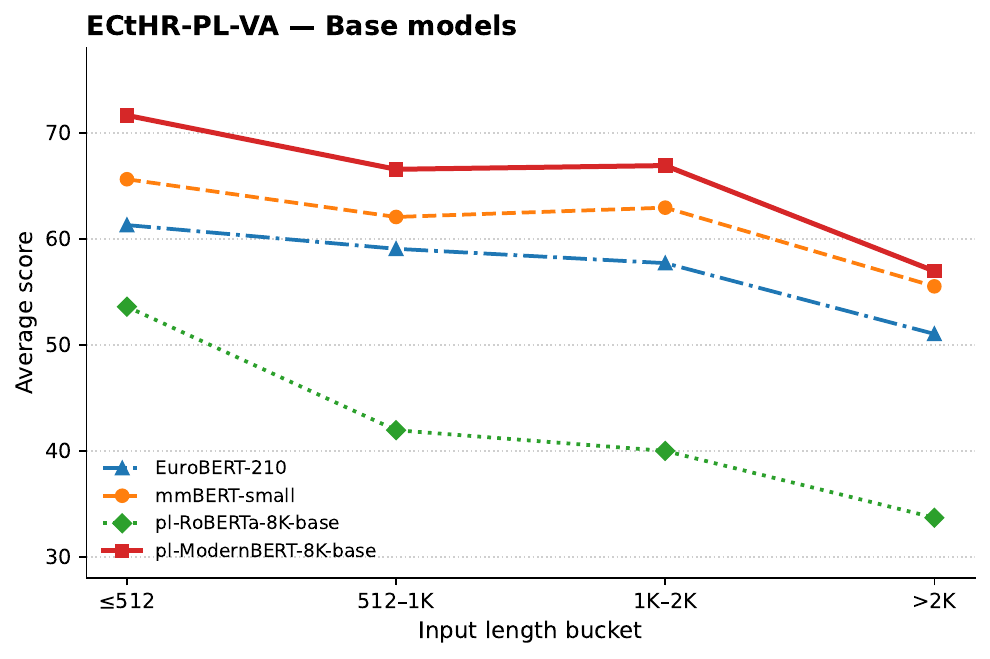}
    \caption{\textsc{ecthr-pl-va}: Base models.}
    \label{fig:length_ecthr_va_base}
\end{subfigure}
\hfill
\begin{subfigure}[t]{0.49\textwidth}
    \centering
    \includegraphics[width=\linewidth]{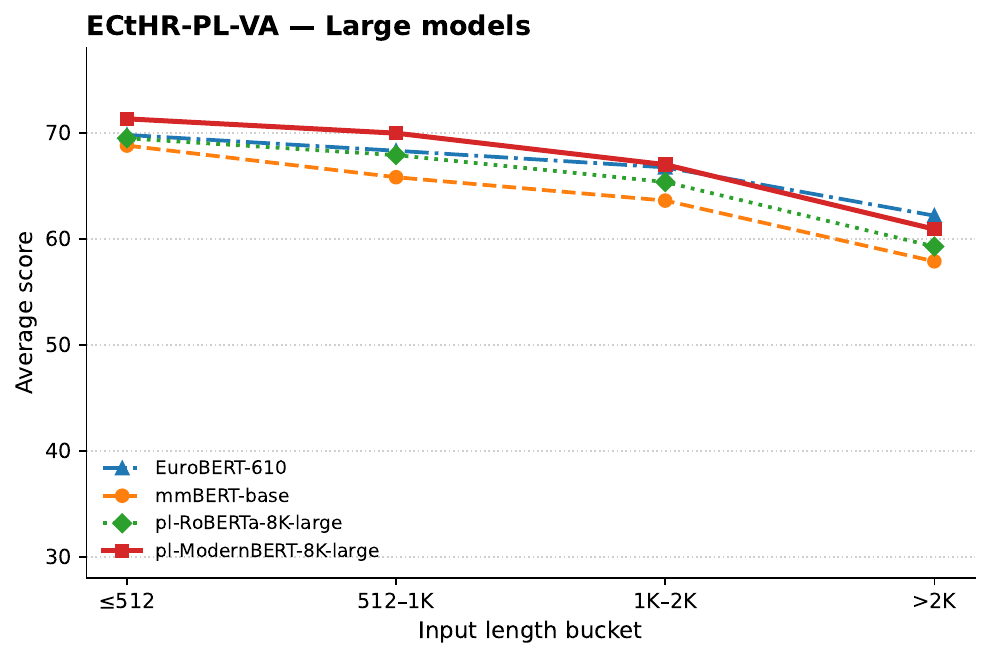}
    \caption{\textsc{ecthr-pl-va}: Large models.}
    \label{fig:length_ecthr_va_large}
\end{subfigure}

\caption{
Performance by input length on the ECtHR-PL-AVA and ECtHR-PL-VA LongContext
tasks. Each point shows the mean test score for examples within the
corresponding length bucket. Results are presented separately for Base and
Large models. Bucket definitions and example counts are provided in
Table~\ref{tab:length_bucket_counts}, and detailed results with variability
across runs are reported in Table~\ref{tab:length_bucket_results}.
}
\label{fig:length_bucket_ecthr}
\end{figure*}

The SCOTUS tasks use wider ranges because their examples are more evenly
distributed across longer input sequences, whereas BookSummary and the ECtHR
tasks contain a larger proportion of shorter examples.
Table~\ref{tab:length_bucket_counts} reports the number and proportion of test
examples assigned to each bucket.
Figures~\ref{fig:length_bucket_scotus_books}
and~\ref{fig:length_bucket_ecthr} present the corresponding performance
breakdowns for Base- and Large-scale models.

\begin{table}[th]
\centering
\small
\setlength{\tabcolsep}{5pt}
\caption{
Input-length bucket definitions and test-set distributions for the SCOTUS,
BookSummary, and ECtHR-PL LongContext tasks. SCOTUS-Dom and SCOTUS-Dec share
the same input documents, as do ECtHR-PL-AVA and ECtHR-PL-VA; therefore,
example counts are reported jointly for each task pair.
}
\label{tab:length_bucket_counts}
\begin{tabular}{llrr}
\toprule
\textbf{Dataset} & \textbf{Bucket}
& \textbf{N} & \textbf{Share (\%)} \\
\midrule
SCOTUS-Dom/Dec
    & $\leq$2K & 122 & 13.10 \\
    & 2K--4K   & 115 & 12.35 \\
    & 4K--6K   & 154 & 16.54 \\
    & $>$6K    & 540 & 58.00 \\
\midrule
BookSummary
    & $\leq$512 & 151 & 12.10 \\
    & 512--1K   & 614 & 49.20 \\
    & 1K--2K    & 382 & 30.61 \\
    & $>$2K     & 101 & 8.09 \\
\midrule
ECtHR-PL-AVA/VA
    & $\leq$512 & 136 & 16.61 \\
    & 512--1K   & 203 & 24.79 \\
    & 1K--2K    & 289 & 35.29 \\
    & $>$2K     & 191 & 23.32 \\
\bottomrule
\end{tabular}
\end{table}


\begin{table*}[t]
\centering
\small
\caption{
Mean scores and sample standard deviations across five fine-tuning runs
for the principal matched Polish-model comparisons. Aggregate scores are
computed independently for each random seed and then summarized across seeds.
All metrics are reported on a 0--100 scale. Boldface marks the higher mean
within each matched RoBERTa--ModernBERT pair.
}
\label{tab:aggregate_variability}
\setlength{\tabcolsep}{4.2pt}
\begin{tabular}{lccc}
\toprule
\textbf{Model}
& \textbf{Short-context Avg}
& \textbf{LongContext Avg}
& \textbf{Overall Avg} \\
\midrule

\multicolumn{4}{l}{\textit{Base scale}} \\

pl-RoBERTa-v2-base
& $83.77 \pm 0.18$
& $57.66 \pm 1.87$
& $79.42 \pm 0.35$ \\

pl-ModernBERT-base
& $\mathbf{84.96 \pm 0.21}$
& $\mathbf{71.59 \pm 1.74}$
& $\mathbf{82.73 \pm 0.34}$ \\

pl-RoBERTa-8K-base
& $84.85 \pm 0.22$
& $67.47 \pm 1.41$
& $81.95 \pm 0.30$ \\

pl-ModernBERT-8K-base
& $\mathbf{85.36 \pm 0.21}$
& $\mathbf{77.15 \pm 1.38}$
& $\mathbf{83.99 \pm 0.29}$ \\

\midrule

pl-RoBERTa-v2-large
& $\mathbf{86.46 \pm 0.24}$
& $70.49 \pm 1.92$
& $83.80 \pm 0.38$ \\

pl-ModernBERT-large
& $\mathbf{86.46 \pm 0.25}$
& $\mathbf{73.58 \pm 1.81}$
& $\mathbf{84.31 \pm 0.37}$ \\

pl-RoBERTa-8K-large
& $\mathbf{86.69 \pm 0.18}$
& $75.88 \pm 1.60$
& $84.89 \pm 0.31$ \\

pl-ModernBERT-8K-large
& $86.43 \pm 0.20$
& $\mathbf{78.49 \pm 1.76}$
& $\mathbf{85.11 \pm 0.34}$ \\

\bottomrule
\end{tabular}
\end{table*}

\paragraph{Length-dependent trends.}
The bucket-level results show that the relative performance of Polish
ModernBERT is particularly consistent at the Base scale. Descriptively, it
achieves the highest mean score in 18 of the 20 task--bucket comparisons for
Base models, compared with 9 of 20 comparisons at the Large scale. The
advantages at the Large scale are therefore more task-dependent, partly
because the corresponding Polish RoBERTa and EuroBERT baselines are already
strong.

The SCOTUS tasks do not exhibit a monotonic decline with increasing input
length. On SCOTUS-Dom, scores generally increase for the intermediate-length
buckets, and Polish ModernBERT-8K-Base obtains the highest score in three of
the four ranges. The Large-scale results are more mixed: Polish RoBERTa-8K
performs best in the 2K--4K and 4K--6K buckets, while Polish ModernBERT-8K
achieves the strongest result for inputs longer than 6K. On SCOTUS-Dec,
Polish ModernBERT-8K-Base leads in all four buckets. The Large variant leads
in three buckets and is effectively tied with EuroBERT-610 in the 2K--4K
range.

A clearer length-related degradation is observed on BookSummary and the ECtHR
tasks. On BookSummary, Polish ModernBERT-8K-Base leads for inputs up to 2K
tokens, whereas Polish RoBERTa-8K-Base performs best in the longest bucket.
At the Large scale, Polish RoBERTa-8K obtains the highest score in all four
BookSummary ranges, although the difference from Polish ModernBERT-8K remains
small for inputs longer than 2K tokens. On both ECtHR tasks, Polish
ModernBERT-8K-Base provides the strongest results across all length buckets.
The Large-scale comparison is less uniform, but Polish ModernBERT remains
competitive and obtains the best result in the longest ECtHR-PL-AVA bucket.

These patterns indicate that input length alone does not fully explain task
difficulty: the SCOTUS results vary non-monotonically across buckets, whereas
BookSummary and ECtHR show more systematic declines for longer inputs. The
results should also be interpreted together with the bucket sizes reported in
Table~\ref{tab:length_bucket_counts}, particularly for the longest
BookSummary range, which contains only 101 test examples.

\begin{table*}[t]
\centering
\scriptsize
\setlength{\tabcolsep}{2.5pt}
\renewcommand{\arraystretch}{1.05}
\caption{
Performance by input length across all LongContext tasks. Each cell reports
the mean score and standard deviation across the available fine-tuning runs.
Boldface marks the highest mean within each model-scale group for a given
length bucket. Bucket
definitions and example counts are provided in
Table~\ref{tab:length_bucket_counts}.
}
\label{tab:length_bucket_results}

\resizebox{\textwidth}{!}{
\begin{tabular}{
ll
cccc
!{\color{gray!60}\vrule width 0.3pt}
cccc
}
\toprule
\textbf{Task}
& \textbf{Length}
& \multicolumn{4}{c}{
    \textbf{Base scale}
}
& \multicolumn{4}{c}{
    \textbf{Large scale}
} \\
\cmidrule(lr){3-6}
\cmidrule(lr){7-10}
&
&
\shortstack{\textbf{EuroBERT}\\\textbf{210m}}
&
\shortstack{\textbf{mmBERT}\\\textbf{small}}
&
\shortstack{\textbf{pl-RoBERTa}\\\textbf{8K-base}}
&
\shortstack{\textbf{pl-ModernBERT}\\\textbf{8K-base}}
&
\shortstack{\textbf{EuroBERT}\\\textbf{610m}}
&
\shortstack{\textbf{mmBERT}\\\textbf{base}}
&
\shortstack{\textbf{pl-RoBERTa}\\\textbf{8K-large}}
&
\shortstack{\textbf{pl-ModernBERT}\\\textbf{8K-large}}
\\
\midrule

\textsc{scotus-dom}
& $\leq$2K
& $80.99 \pm 2.62$
& $75.90 \pm 4.94$
& $75.74 \pm 3.21$
& $\mathbf{81.15 \pm 2.39}$
& $\mathbf{82.79 \pm 2.15}$ 
& $80.17 \pm 1.87$
& $79.67 \pm 2.04$
& $80.33 \pm 2.17$
\\

& 2K--4K
& $\mathbf{83.31 \pm 2.17}$
& $77.22 \pm 2.71$
& $77.74 \pm 1.80$
& $80.35 \pm 2.65$
& $79.13 \pm 2.25$  
& $79.30 \pm 2.25$
& $\mathbf{84.70 \pm 2.58}$
& $80.58 \pm 0.50$
\\

& 4K--6K
& $85.97 \pm 2.54$
& $80.78 \pm 3.24$
& $84.03 \pm 1.34$
& $\mathbf{87.01 \pm 1.78}$
& $85.06 \pm 3.15$
& $84.80 \pm 3.86$
& $\mathbf{90.65 \pm 0.87}$
& $85.28 \pm 0.99$
\\

& $>$6K
& $84.41 \pm 1.18$
& $79.18 \pm 0.81$
& $79.44 \pm 0.41$
& $\mathbf{85.37 \pm 0.73}$
& $85.37 \pm 1.75$
& $83.59 \pm 1.23$
& $84.41 \pm 0.62$
& $\mathbf{86.30 \pm 1.03}$
\\

\midrule

\textsc{scotus-dec}
& $\leq$2K
& $76.72 \pm 1.70$
& $73.16 \pm 2.16$
& $67.76 \pm 4.94$
& $\mathbf{79.78 \pm 4.21}$
& $78.14 \pm 3.31$
& $76.89 \pm 1.96$
& $75.00 \pm 1.42$
& $\mathbf{81.64 \pm 5.46}$
\\

& 2K--4K
& $75.83 \pm 3.95$
& $61.96 \pm 2.60$
& $62.32 \pm 1.00$
& $\mathbf{77.10 \pm 2.80}$
& $\mathbf{78.84 \pm 3.62}$
& $70.43 \pm 3.89$
& $70.22 \pm 2.05$
& $78.78 \pm 6.42$
\\

& 4K--6K
& $73.12 \pm 2.37$
& $56.65 \pm 1.34$
& $63.20 \pm 3.07$
& $\mathbf{74.68 \pm 1.30}$
& $69.48 \pm 1.30$
& $65.84 \pm 4.70$
& $63.96 \pm 3.29$
& $\mathbf{72.47 \pm 4.94}$
\\

& $>$6K
& $77.11 \pm 2.19$
& $59.72 \pm 0.49$
& $63.40 \pm 2.71$
& $\mathbf{77.47 \pm 3.85}$
& $77.22 \pm 2.67$
& $71.22 \pm 2.31$
& $67.36 \pm 1.50$
& $\mathbf{78.97 \pm 6.48}$
\\

\midrule

\textsc{booksummary}
& $\leq$512
& $82.94 \pm 5.93$
& $84.24 \pm 0.56$
& $90.90 \pm 0.83$
& $\mathbf{92.98 \pm 1.73}$
& $93.64 \pm 1.37$
& $89.67 \pm 2.91$
& $\mathbf{93.91 \pm 0.86}$
& $93.25 \pm 1.78$
\\

& 512--1K
& $84.00 \pm 4.28$
& $85.08 \pm 0.45$
& $90.47 \pm 0.60$
& $\mathbf{92.08 \pm 0.42}$
& $93.78 \pm 0.89$
& $89.19 \pm 1.57$
& $\mathbf{95.18 \pm 0.84}$
& $93.03 \pm 0.67$
\\

& 1K--2K
& $80.04 \pm 4.70$
& $82.51 \pm 1.02$
& $87.50 \pm 1.44$
& $\mathbf{88.48 \pm 0.49}$
& $90.47 \pm 1.80$
& $86.75 \pm 1.26$
& $\mathbf{91.52 \pm 1.24}$
& $89.22 \pm 1.16$
\\

& $>$2K
& $79.21 \pm 1.40$
& $65.15 \pm 4.34$
& $\mathbf{82.43 \pm 1.69}$
& $80.40 \pm 1.90$
& $83.37 \pm 2.66$
& $74.46 \pm 3.00$
& $\mathbf{85.35 \pm 2.26}$
& $84.95 \pm 3.98$
\\

\midrule

\textsc{ecthr-pl-ava}
& $\leq$512
& $68.32 \pm 0.49$
& $62.68 \pm 2.93$
& $67.98 \pm 1.37$
& $\mathbf{75.16 \pm 1.39}$
& $72.01 \pm 1.60$
& $67.52 \pm 1.26$
& $\mathbf{74.02 \pm 1.40}$
& $72.42 \pm 0.23$
\\

& 512--1K
& $63.28 \pm 1.56$
& $59.72 \pm 3.26$
& $66.91 \pm 0.66$
& $\mathbf{68.13 \pm 1.40}$
& $68.57 \pm 1.82$
& $63.28 \pm 2.68$
& $69.17 \pm 2.17$
& $\mathbf{71.63 \pm 2.15}$
\\

& 1K--2K
& $62.90 \pm 1.26$
& $61.00 \pm 1.09$
& $64.34 \pm 1.21$
& $\mathbf{67.99 \pm 0.72}$
& $\mathbf{69.81 \pm 1.29}$
& $64.13 \pm 1.27$
& $67.56 \pm 0.57$
& $69.32 \pm 1.01$
\\

& $>$2K
& $57.19 \pm 2.00$
& $56.97 \pm 1.39$
& $60.35 \pm 0.50$
& $\mathbf{64.28 \pm 0.64}$
& $63.58 \pm 1.21$
& $59.34 \pm 1.79$
& $64.36 \pm 0.52$
& $\mathbf{65.33 \pm 1.51}$
\\

\midrule

\textsc{ecthr-pl-va}
& $\leq$512
& $61.31 \pm 3.07$
& $65.63 \pm 0.80$
& $53.60 \pm 3.86$
& $\mathbf{71.65 \pm 2.14}$
& $69.81 \pm 2.67$
& $68.80 \pm 4.07$
& $69.49 \pm 3.27$
& $\mathbf{71.33 \pm 3.16}$
\\

& 512--1K
& $59.07 \pm 1.73$
& $62.07 \pm 2.52$
& $41.96 \pm 3.44$
& $\mathbf{66.57 \pm 1.77}$
& $68.31 \pm 2.83$
& $65.82 \pm 3.29$
& $67.91 \pm 0.98$
& $\mathbf{69.97 \pm 1.99}$
\\

& 1K--2K
& $57.71 \pm 2.25$
& $62.95 \pm 1.37$
& $40.00 \pm 1.61$
& $\mathbf{66.92 \pm 1.16}$
& $66.75 \pm 2.23$
& $63.61 \pm 1.51$
& $65.37 \pm 0.97$
& $\mathbf{67.01 \pm 1.50}$
\\

& $>$2K
& $51.04 \pm 3.52$
& $55.52 \pm 2.28$
& $33.70 \pm 1.72$
& $\mathbf{56.96 \pm 1.48}$
& $\mathbf{62.19 \pm 1.48}$
& $57.88 \pm 2.37$
& $59.28 \pm 2.59$
& $60.92 \pm 1.56$
\\

\bottomrule
\end{tabular}
}
\end{table*}

\raggedbottom
\section{Efficiency Analysis}
\label{app:efficiency}

We provide additional efficiency measurements for all evaluated encoder
models. For each model, we report peak GPU memory usage, per-sample latency,
and the corresponding average task score.

\paragraph{Measurement setup.}
All measurements were conducted on a single NVIDIA H100 GPU using BF16
precision. We used a batch size of 256 for the 512-token measurements on
POLEMO2.0-IN and a batch size of 64 for the long-context measurements on
SCOTUS-Dom. Each measurement was repeated five times, and we report the mean
throughput and peak GPU memory usage across repetitions. GPU operations were
synchronized before and after each timed region. Peak memory usage was
measured using \texttt{torch.cuda.max\_memory\_allocated()} after resetting
the peak-memory statistics. The timed region covered model inference and
excluded tokenization. Within each setting, all compared models were
evaluated on the same examples under an otherwise identical inference
configuration.

\begin{figure*}[t]
    \centering
    \includegraphics[width=\textwidth]{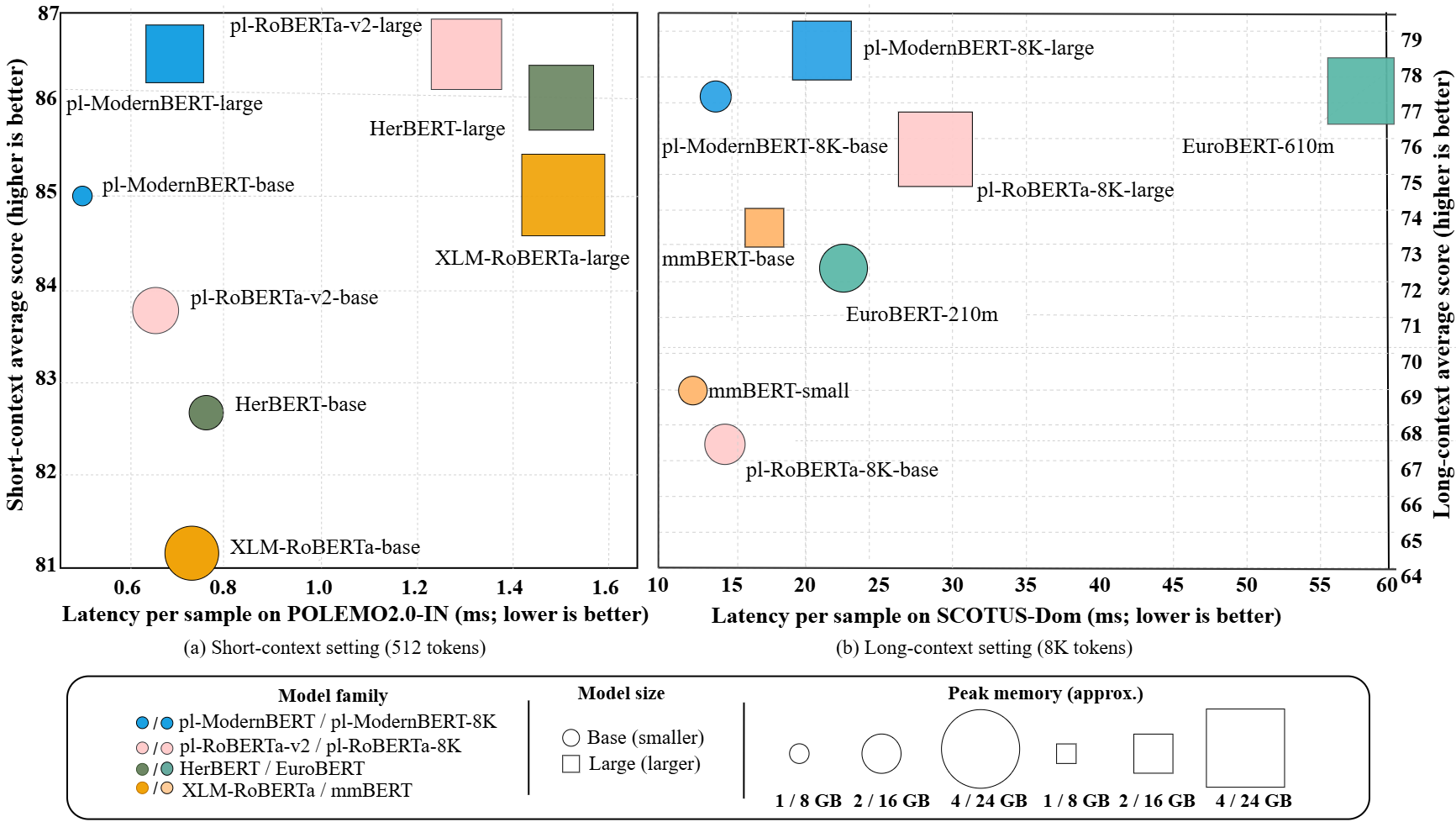}
   \caption{
Quality--efficiency trade-offs for the evaluated encoder models on Polish
tasks. The x-axis shows
latency per sample (lower is better), the y-axis shows the average task score
(higher is better), marker size reflects peak memory usage, and marker shape
distinguishes Base and Large variants. Panel (a) presents short-context
results, while panel (b) shows long-context results. Marker sizes are scaled separately for the two panels; slash-separated
model-family names and memory labels correspond to panels (a) and (b),
respectively.
}
    \label{fig:tradeoff-main}
\end{figure*}

Per-sample latency was derived as the inverse of throughput and expressed in
milliseconds per sample. The 512-token measurements are based on
POLEMO2.0-IN, whereas the long-context measurements are based on SCOTUS-Dom.
Tables~\ref{tab:efficiency-short} and~\ref{tab:efficiency-long} provide the
numerical results, while Figure~\ref{fig:tradeoff-main} visualizes the
quality--efficiency trade-offs.

\begin{table*}[t]
\centering
\small
\caption{
Efficiency measurements for 512-token encoder models.
Latency denotes per-sample latency derived from throughput on POLEMO2.0-IN,
and memory denotes the observed peak memory usage during inference.
The final column reports the average over the short-context tasks.
}
\label{tab:efficiency-short}
\begin{tabular}{lrrr}
\toprule
\textbf{Model} & \textbf{Memory [MB]} & \textbf{Latency [ms/sample]} & \textbf{Short-context avg.} \\
\midrule
XLM-RoBERTa-base      & 2,764 & 0.71 & 81.19 \\
HerBERT-base          & 1,784 & 0.74 & 82.69 \\
pl-RoBERTa-v2-base    & 2,344 & 0.65 & 83.77 \\
pl-ModernBERT-base    & 1,084 & 0.48 & 84.96 \\
\midrule
XLM-RoBERTa-large     & 4,144 & 1.50 & 84.98 \\
HerBERT-large         & 3,304 & 1.49 & 85.95 \\
pl-RoBERTa-v2-large   & 3,604 & 1.29 & 86.46 \\
pl-ModernBERT-large   & 2,946 & 0.69 & 86.46 \\
\bottomrule
\end{tabular}
\end{table*}

\begin{table*}[t]
\centering
\small
\caption{
Efficiency measurements for long-context encoder models.
Latency denotes per-sample latency derived from throughput on SCOTUS-Dom,
and memory denotes the observed peak GPU memory usage during inference.
The final column reports the average over the five LongContext tasks.
}
\label{tab:efficiency-long}
\begin{tabular}{lrrr}
\toprule
\textbf{Model} & \textbf{Memory [MB]} &
\textbf{Latency [ms/sample]} & \textbf{LongContext avg.} \\
\midrule
EuroBERT-210m               & 14,286 & 22.57 & 72.30 \\
mmBERT-small                &  8,546 & 12.12 & 69.02 \\
pl-RoBERTa-8K-base          & 12,208 & 14.91 & 67.47 \\
pl-ModernBERT-8K-base       &  9,250 & 13.98 & 77.15 \\
\midrule
EuroBERT-610m               & 19,670 & 57.63 & 77.36 \\
mmBERT-base                 & 11,500 & 17.04 & 73.44 \\
pl-RoBERTa-8K-large         & 22,166 & 28.96 & 75.88 \\
pl-ModernBERT-8K-large      & 17,554 & 21.69 & 78.49 \\
\bottomrule
\end{tabular}
\end{table*}

\paragraph{Short-context models.}
In the 512-token setting, pl-ModernBERT-base achieves the lowest latency and
peak memory usage among all evaluated short-context models, while also
obtaining the highest average score among the base-size models. Compared with
pl-RoBERTa-v2-base, it improves the average score from 83.77 to 84.96, reduces
latency from 0.65 to 0.48 ms per sample, and lowers peak memory usage from
2,344 to 1,084 MB.

At the Large scale, pl-ModernBERT-large matches the average score of
pl-RoBERTa-v2-large at 86.46, while reducing latency from 1.29 to 0.69 ms per
sample and peak memory usage from 3,604 to 2,946 MB. This corresponds to
approximately 47\% lower latency, or equivalently about 1.9$\times$ higher
throughput, while preserving the same average task performance.

\paragraph{Long-context models.}
The long-context results show a similar trend.
pl-ModernBERT-8K-base reaches a long-context average score of 77.15,
outperforming pl-RoBERTa-8K-base by 9.68 points, while also using less
memory and slightly lower latency. Among larger long-context models,
pl-ModernBERT-8K-large obtains the highest long-context average score of
78.49. Compared with pl-RoBERTa-8K-large, it improves the score by
2.61 points, reduces latency from 28.97 ms to 21.69 ms per sample, and
decreases peak memory usage from 22,166 MB to 17,554 MB.

\section{Retrieval experiments}
\label{app:retrieval}
One of the common applications of encoder-only models is dense text retrieval. We therefore evaluate how Polish ModernBERT performs in this setting relative to other encoders supporting Polish. We first apply contrastive fine-tuning to all models using a training corpus composed of Polish retrieval datasets. We then evaluate the resulting models on PIRB \citep{dadas-etal-2024-pirb}, a comprehensive Polish retrieval benchmark spanning 41 datasets. PIRB consolidates previously introduced evaluation suites, including BEIR-PL \citep{wojtasik2024beir}, MaupQA \citep{rybak2023maupqa}, and PolEval-2022 retrieval challenge \citep{kobylinski2023poleval}, and extends them with 10 new datasets.

\begin{table}[t]
\centering
\small
\setlength{\tabcolsep}{5pt}
\caption{
Mean NDCG@10 on PIRB for dedicated multilingual embedding models.
}
\label{tab:retrieval_dedicated}
\begin{tabular}{lr}
\toprule
\textbf{Model} & \textbf{NDCG@10} \\
\midrule
Qwen3-Embedding-0.6B (596M)            & 50.71 \\
BGE-M3 (568M)                          & 55.98 \\
Jina-Embeddings-V5-Text-Small (677M)   & 57.19 \\
Multilingual-E5-Large (560M)           & 57.29 \\
Snowflake-Arctic-Embed-L-v2.0 (568M)   & 59.22 \\
\bottomrule
\end{tabular}
\end{table}

Because PIRB includes datasets containing long documents, we restricted the comparison to models supporting a context length of 8K tokens. The fine-tuning corpus comprised 12 datasets. Nine of these were training splits from datasets included in Polish BEIR \citep{wojtasik2024beir}: MS MARCO, SciFact, SciDocs, Natural Questions, NFCorpus, HotpotQA, FiQA, FEVER, and ArguAna. We additionally included Polish translations of ELI5 \citep{fan2019eli5} and GooAQ \citep{khashabi2021gooaq}, as well as the Polish ZnanyLekarz dataset \citep{dadas2024assessing}. In total, the corpus contained approximately 4.5 million queries and more than 15 million passages. We fine-tuned each model for 10 epochs using contrastive learning with in-batch negatives and a batch size of 1024. We used a learning rate scheduler with 200 warmup steps, a maximum learning rate of $2\times10^{-6}$, and linear decay.

For each model, we report the mean NDCG@10 score across all 41 retrieval
datasets. Table~\ref{tab:retrieval_results} presents the encoders fine-tuned
using our shared contrastive-learning protocol, divided into two size groups
using 300M parameters as the cutoff.
Table~\ref{tab:retrieval_dedicated} provides additional context through
external PIRB results for dedicated multilingual embedding models, including
Qwen3-Embedding \citep{zhang2025qwen3}, BGE-M3 \citep{chen2024m3},
Jina-Embeddings-V5 \citep{akram2026jina}, Multilingual-E5
\citep{yu2025arctic}, and Snowflake-Arctic
\citep{wang2024multilingual}.

Under the shared fine-tuning protocol, the monolingual Polish encoders
outperform the multilingual mmBERT and EuroBERT models in both size groups.
Among the smaller models, Polish ModernBERT-8K-Base achieves the highest
score, outperforming Polish RoBERTa-8K-Base by 1.26 NDCG@10 points
(55.36 vs.\ 54.10). At the larger scale, the two Polish encoders obtain
comparable results, with Polish RoBERTa-8K-Large slightly outperforming
Polish ModernBERT-8K-Large by 0.26 points (58.42 vs.\ 58.16).

The fine-tuned Polish encoders also compare favorably with dedicated
multilingual embedding models. The Polish models were adapted using only the
shared contrastive-learning procedure, whereas dedicated embedding models may
rely on more elaborate multi-stage pipelines involving retrieval-oriented
pretraining, knowledge distillation, or additional training data.
Nevertheless, both larger Polish encoders outperform all dedicated embedding
baselines included in the comparison except
Snowflake-Arctic-Embed-L-v2.0, which achieves the highest overall score of
59.22.

\subsection{Reference Embedding Models}
\label{app:retrieval_embedding_references}

We additionally report dedicated multilingual embedding models as external
reference points for the PIRB score range. These models are not treated as
directly comparable baselines because they differ from the shared-protocol
encoders in training data, objectives, and evaluation setup. As shown in
Table~\ref{tab:retrieval_dedicated}, their scores range from 50.71 to 59.22
NDCG@10, with Snowflake-Arctic-Embed-L-v2.0 obtaining the highest result.

\subsection{BookSummary Claim-Generation Prompts}
\label{app:booksummary_prompts}

The BookSummary task was constructed using a two-stage generation procedure.
First, GLM-4.6 generated a short claim supported by the corresponding plot
summary, with an explicit preference for information appearing in the second
half of the document. Second, the model transformed the supported claim into
a closely matched but factually inconsistent claim. Because the documents and
generated claims were in Polish, both prompts were executed in Polish. English
translations are provided below for readability, followed by the original
Polish prompts used during dataset construction. The English versions were not
used for generation. The placeholders \texttt{\{\{doc\}\}} and
\texttt{\{\{claim\}\}} denote the plot summary and the previously generated
supported claim, respectively.

\subsubsection{Supported-Claim Generation}

\paragraph{English translation.}

\begin{quote}
\small
\textbf{System message.}

Based on the document below, generate one short claim consisting of 2--3
sentences. Requirements:
\begin{itemize}
    \item The claim must be consistent with the content of the document.
    \item The claim should concern one or two facts from the document.
    \item The claim should consist of 2--3 sentences.
    \item The claim must be unambiguous and verifiable solely on the basis of
    the document.
    \item Do not use external knowledge.
    \item Do not quote the document verbatim.
    \item The claim should describe important information from the document.
    \item Whenever possible, focus the claim on facts appearing in the second
    half of the document.
    \item Return only the claim, without any additional comments.
\end{itemize}

\medskip
\textbf{User message.}

\texttt{\{\{doc\}\}}
\end{quote}

\paragraph{Original Polish prompt.}

\begin{quote}
\small
\textbf{System message.}

Na podstawie poniższego dokumentu wygeneruj jedno krótkie, 2--3 zdaniowe
twierdzenie. Wymagania:
\begin{itemize}
    \item Twierdzenie musi być zgodne z treścią dokumentu.
    \item Twierdzenie powinno dotyczyć jednego lub dwóch faktów z artykułu.
    \item Twierdzenie powinno mieć długość 2 lub 3 zdań.
    \item Twierdzenie musi być jednoznaczne i możliwe do zweryfikowania
    wyłącznie na podstawie dokumentu.
    \item Nie należy używać wiedzy zewnętrznej.
    \item Nie należy cytować dokumentu dosłownie.
    \item Twierdzenie powinno opisywać istotną informację z dokumentu.
    \item Postaraj się, jeśli to możliwe, żeby twierdzenie skupiało się na
    faktach z drugiej połowy dokumentu.
    \item Zwróć wyłącznie treść twierdzenia bez dodatkowych komentarzy.
\end{itemize}

\medskip
\textbf{User message.}

\texttt{\{\{doc\}\}}
\end{quote}

\subsubsection{Unsupported-Claim Generation}

\paragraph{English translation.}

\begin{quote}
\small
\textbf{System message.}
Based on the document and the provided claim, generate a new false claim that
is highly similar to the original one but inconsistent with the content of
the document. Requirements:
\begin{itemize}
    \item Preserve a similar length, style, and structure.
    \item The claim should still consist of 2--3 sentences.
    \item The factual inconsistency should be easy to verify using the
    passages of the document to which the supported claim refers.
    \item The inconsistency should not be limited to a single fact, but should
    more broadly contradict the facts stated in the original claim.
    \item Do not introduce new information unrelated to the document.
    \item Do not explain the introduced changes.
\end{itemize}
Return only the new claim.

\medskip
\textbf{User message.}

Document: \texttt{\{\{doc\}\}}

Original claim: \texttt{\{\{claim\}\}}
\end{quote}

\paragraph{Original Polish prompt.}

\begin{quote}
\small
\textbf{System message.}
Na podstawie dokumentu oraz podanego twierdzenia wygeneruj nowe, fałszywe
twierdzenie, które będzie bardzo podobne do oryginalnego, ale będzie niezgodne
z treścią dokumentu. Wymagania:
\begin{itemize}
    \item Zachowaj podobną długość, styl i strukturę.
    \item Twierdzenie powinno nadal składać się z 2--3 zdań.
    \item Fałszywość twierdzenia powinna być łatwo zweryfikowana na podstawie
    fragmentów treści dokumentu, do których odnosi się prawdziwe twierdzenie.
    \item Fałszywość nie powinna dotyczyć jednego faktu, ale raczej powinna
    szerzej zaprzeczać faktom w oryginalnym twierdzeniu.
    \item Nie dodawaj nowych informacji, które nie są związane z dokumentem.
    \item Nie wyjaśniaj wprowadzonych zmian.
\end{itemize}
Zwróć wyłącznie nowe twierdzenie.

\medskip
\textbf{User message.}

Dokument: \texttt{\{\{doc\}\}}

Oryginalne twierdzenie: \texttt{\{\{claim\}\}}
\end{quote}

\end{document}